\documentclass[letterpaper]{article} 
\usepackage{aaai23}  
\usepackage{times}  
\usepackage{helvet}  
\usepackage{courier}  
\usepackage[hyphens]{url}  
\usepackage{graphicx} 
\usepackage{natbib}  
\usepackage{caption} 
\usepackage{algorithm}
\usepackage{algorithmic}
\usepackage{pifont}
\usepackage{color, colortbl}
\definecolor{Gray}{gray}{0.9}
\usepackage{lipsum}
\usepackage{algorithmic}
\usepackage{xcolor}
\usepackage{multirow}
\usepackage{hyperref}
\usepackage{url}
\usepackage{newfloat}
\usepackage{listings}
\usepackage{newfloat}
\usepackage{listings}
\DeclareCaptionStyle{ruled}{labelfont=normalfont,labelsep=colon,strut=off} 
\floatstyle{ruled}
\newfloat{listing}{tb}{lst}{}
\floatname{listing}{Listing}
\title{Unbiased Heterogeneous Scene Graph Generation with Relation-aware\\ Message Passing Neural Network}
\author{
    Kanghoon Yoon\textsuperscript{\rm 1}\equalcontrib,
    Kibum Kim\textsuperscript{\rm 1}\equalcontrib,
    Jinyoung Moon\textsuperscript{\rm 3,4},
    Chanyoung Park\textsuperscript{\rm 1,2}\thanks{Corresponding author.}
}
\affiliations{
    \textsuperscript{\rm 1}Dept. of Industrial and Systems Engineering, KAIST, Daejeon, Republic of Korea\\
    \textsuperscript{\rm 2}Graduate School of Artificial Intelligence, KAIST, Daejeon, Republic of Korea\\
    \textsuperscript{\rm 3}Electronics and Telecommunications Research Institute, 218 Gajeong-ro, Yuseong-gu, Daejeon, Republic of Korea\\
    \textsuperscript{\rm 4} ETRI School, University of Science and Technology, 218 Gajeong-ro, Yuseong-gu, Daejeon, Republic of Korea\\    
    ykhoon08@kaist.ac.kr, kb.kim@kaist.ac.kr, jymoon@etri.re.kr,
    cy.park@kaist.ac.kr
}
\usepackage{bibentry}
\usepackage{amsmath, amssymb}
\newcommand{\proposed}{\textsf{\footnotesize HetSGG}}

\begin{document}
\maketitle
\begin{abstract}
 Recent scene graph generation (SGG) frameworks have focused on learning complex relationships among multiple objects in an image. Thanks to the nature of the message passing neural network (MPNN) that models high-order interactions between objects and their neighboring objects, they are dominant representation learning modules for SGG. However, existing MPNN-based frameworks assume the scene graph as a {homogeneous} graph, which restricts the context-awareness of visual 
 relations between objects. That is, 
 they overlook the fact that the relations tend to be highly dependent on the objects with which the relations are associated. 
 In this paper, we propose an {unbiased} heterogeneous scene graph generation (\proposed) framework that captures relation-aware context using message passing neural networks. We devise a novel message passing layer, called relation-aware message passing neural network (RMP), that aggregates the contextual information of an image considering the predicate type between objects.
 Our extensive evaluations demonstrate that~\proposed~outperforms state-of-the-art methods, especially outperforming on tail predicate classes. The source code for~\proposed~is available at \url{https://github.com/KanghoonYoon/hetsgg-torch}.
\end{abstract}

\section{Introduction}
Scene graph generation (SGG) is a fundamental visual understanding task, which aims to identify objects from an image and detect their relations (i.e., predicate\footnote{We use ``relation'' and ``predicate'' interchangeably.}), which can be represented in a triplet format: $\langle$\textsf{subject}, \textsf{predicate}, \textsf{object}$\rangle$. A compact structural scene representation is beneficial to various image applications such as visual question answering~\cite{vqa_2,vqa_3}, image captioning~\cite{image_caption_3}, and image retrieval~\cite{image_retrieval_3,image_retrieval_4}. 
Hence, recent years have seen significant progress in developing methods for SGG.

\begin{figure}[t]
    \begin{center}
        \includegraphics[width=0.46\textwidth]{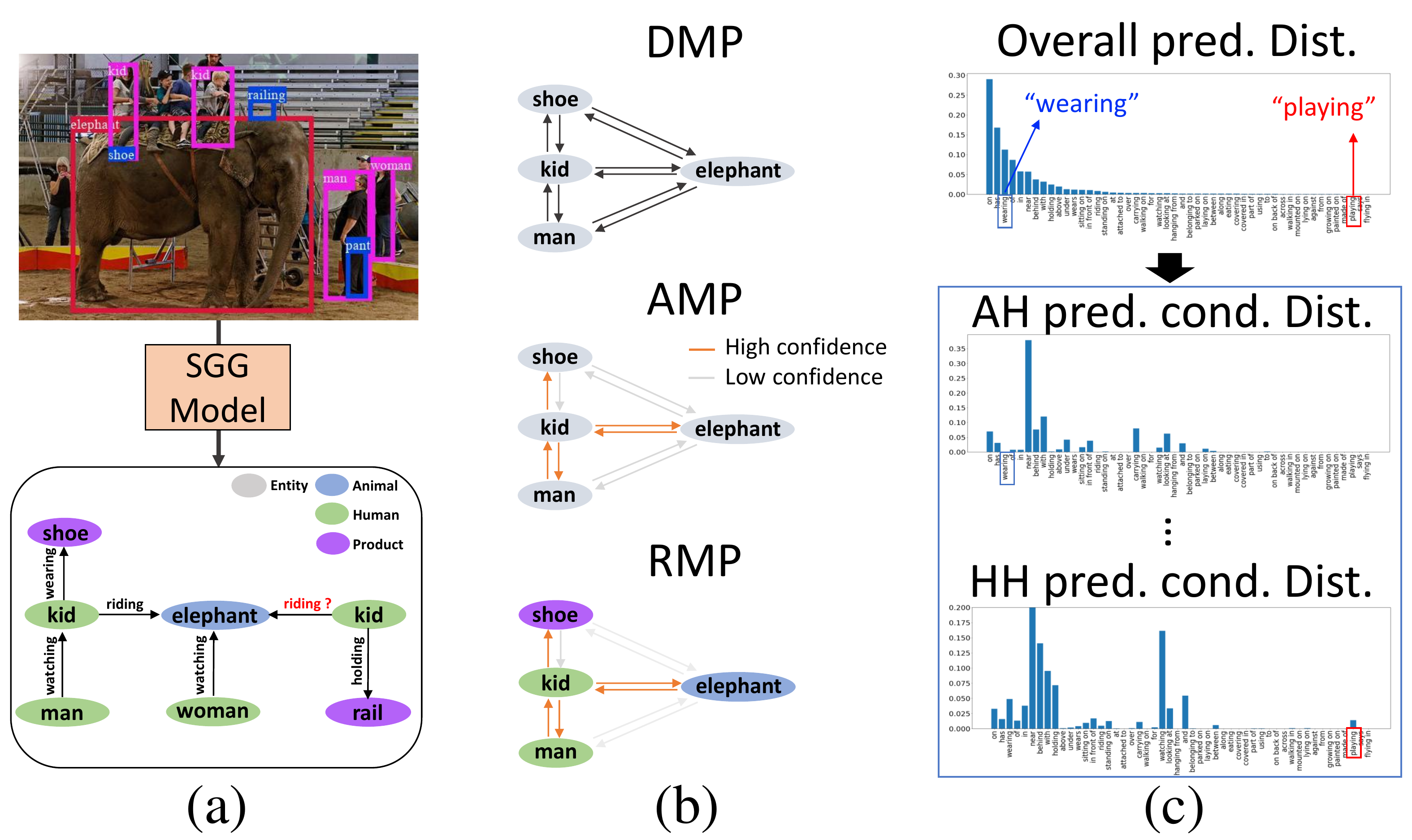}
    \end{center}
    \caption{\textbf{(a)}: Scene graph represented as a heterogeneous graph.
    \textbf{(b)} Comparisons of various MPNNs. \textbf{(c)} Overall predicate distribution and predicate distributions conditioned on certain predicate types.}
    \label{fig:fig_1}
    \vspace{-1ex}
\end{figure}

\looseness=-2
However, recent SGG methods usually have shown unsatisfactory performance due to the difficulty of learning complex relationships among multiple objects in an image. 
An effective scene graph representation should include the contextual information of an image. For intuitively understanding what it means by contextual information,  consider the triplet $\langle$\textsf{kid}, \textsf{riding}, \textsf{elephant}$\rangle$ in Figure~\ref{fig:fig_1}(a). Understanding the context around this triplet, such as ``\textsf{kid holding rail}'' and ``\textsf{woman watching elephant}'', would be helpful for predicting the \textsf{riding} predicate between \textsf{kid} and \textsf{elephant}, compared with the case when \textsf{kid} and \textsf{elephant} are considered independently. 
Hence, recent studies for SGG mainly focus on capturing the contextual information based on message passing neural networks (MPNNs). Thanks to the nature of the MPNNs that model high-order interactions between objects and their neighboring objects, they can naturally capture the visual context of an object from nearby objects that exist together. Specifically, Graph R-CNN~\cite{graph_rcnn} designs an MPNN to learn filtered contextual information from neighbors by identifying important objects and relations. Moreover, direction-aware message passing neural network (DMP) \cite{gps_net} considers the direction to which the messages are propagated between two objects, and adaptive message passing neural network (AMP) \cite{bgnn} prevents irrelevant proposal pairs from interacting with each other based on the confidence of interactions (Figure~\ref{fig:fig_1}(b)). 
In summary, recent SGG methods mainly focus on designing new MPNN architectures aiming at capturing the contextual information of an image, thereby increasing the 
context-awareness of visual relations between objects.

\looseness=-2
Although the aforementioned existing methods for SGG show effectiveness in understanding the visual context by using advanced MPNNs, they commonly consider the scene graph as a \textit{homogeneous} graph, which in turn restricts the 
context-awareness of the visual relations between objects.
A {homogeneous} graph considers all its nodes (i.e., objects) and edges (i.e., predicates) to be of a single type.
For this reason, existing MPNN-based methods overlook the fact that predicates tend to be highly dependent on the objects with which the predicates are associated. 
For example, consider the triplet $\langle$\textsf{kid}, \textsf{riding}, \textsf{elephant}$\rangle$ in Figure~\ref{fig:fig_1}(a). Although a \textsf{kid} can ride an \textsf{elephant}, the opposite direction is unlikely to happen, 
i.e., an \textsf{elephant} usually does not ride a \textsf{kid}
because it is usually ``Human'' that rides ``Animal.''
However, as existing MPNN-based methods consider the scene graph as a homogeneous graph, ``Human''-typed objects and ``Animal''-typed objects cannot be distinguished, which eventually fails to explicitly capture such dependencies. 

In this paper, we propose an \textit{unbiased} {heterogeneous scene graph} generation (\proposed) framework that captures \textit{relation-aware context}.
The main idea is to treat each relation differently according to its type.
More precisely, we
devise a novel message passing layer, called relation-aware message passing neural network (RMP), that aggregates the contextual information of an image considering the predicate type between objects, where the predicate type is determined by the associated object types. 
For example, given a triplet $\langle$\textsf{subject}, \textsf{predicate}, \textsf{object}$\rangle$, 
if \textsf{subject} and \textsf{object} are assigned ``Human (H)'' and ``Animal (A)'' types, respectively,
then the type of \textsf{predicate} is ``Human-Animal (HA).'' More precisely, we first construct a heterogeneous graph
based on the objects detected by an off-the-shelf object detector (e.g., Faster R-CNN~\cite{faster_rcnn}). Then, RMP propagates intra- and inter-relation messages with an attention mechanism to learn the relation-aware context. As described in Figure~\ref{fig:fig_1}(b), 
RMP is the only MPNN layer that fully utilizes the heterogeneity of a scene graph thereby capturing the semantics of the relations. 
It is important to note that RMP is a general framework that subsumes existing mainstream MPNN-based SGG methods, i.e., DMP and AMP. Specifically, RMP is direction-aware by its design, because the predicate types already contain the directional information (e.g., RMP treats $\textsf{kid} \rightarrow \textsf{elephant}$ and $\textsf{elephant} \rightarrow \textsf{kid}$ differently by assigning each relation ``HA'' and ``AH'' predicate type, respectively). Besides, RMP is a generalized version of AMP in that the attention function of RMP is predicate type-aware, whereas that of AMP assumes the predicates are of a single type.

Moreover, considering a scene graph as a heterogeneous graph naturally alleviates the biased prediction problem incurred by the long-tail predicate class distribution. Figure~\ref{fig:fig_1}(c) presents the overall predicate distribution, and the predicate type conditional distributions when considering only the ``AH'' and ``HH'' predicate types. We observe that head predicates in the overall predicate distribution are not anymore considered as head predicates in the ``AH'' or ``HH'' predicate type conditional distributions.
For example, \textsf{wearing} appears at the top-3
in the overall distribution, whereas it rather belongs to a tail predicate class in the ``AH'' predicate type conditional distribution. On the other hand, \textsf{playing} rarely appears in the overall distribution, whereas its
proportion increases in the ``HH'' predicate type conditional distribution. 
This implies that each predicate type exhibits different head/body/tail predicate class distribution.
In this regard, since RMP independently processes all the predicate types, which is the followed by an aggregation,~\proposed~naturally relieves the bias of the final prediction model.
Although existing methods for unbiased prediction \cite{bgnn,gps_net,unbiased,chen2022resistance,Desai_2021_ICCV} improve performance on tail predicates, the performance on head predicates is rather sacrificed. On the other hand,~\proposed~greatly improves the performance on tail predicates, \textit{while maintaining competitive performance on head predicates}.

Our contributions are summarized as follows:
\begin{itemize}
    \item To the best of our knowledge,~\proposed~is the first work to reformulate the SGG task, in which the scene graph was considered as a homogeneous graph, in the light of a heterogeneous graph.
    \item \proposed~is model-agnostic in that it can be adopted to any MPNN-based SGG methods. In this work, we adopt~\proposed~to two recent SGG methods, i.e., Graph R-CNN~\cite{graph_rcnn} and BGNN~\cite{bgnn}, and demonstrate that~\proposed~further improves upon them.
    \item Through extensive experiments on Visual Genome and Open Images, we demonstrate that~\proposed~is superior to state-of-the-art baselines, and that the performance on tail predicate classes in particular improves greatly.
\end{itemize}

\begin{figure*}[t]
    \begin{center}
        \includegraphics[width=.73\textwidth]{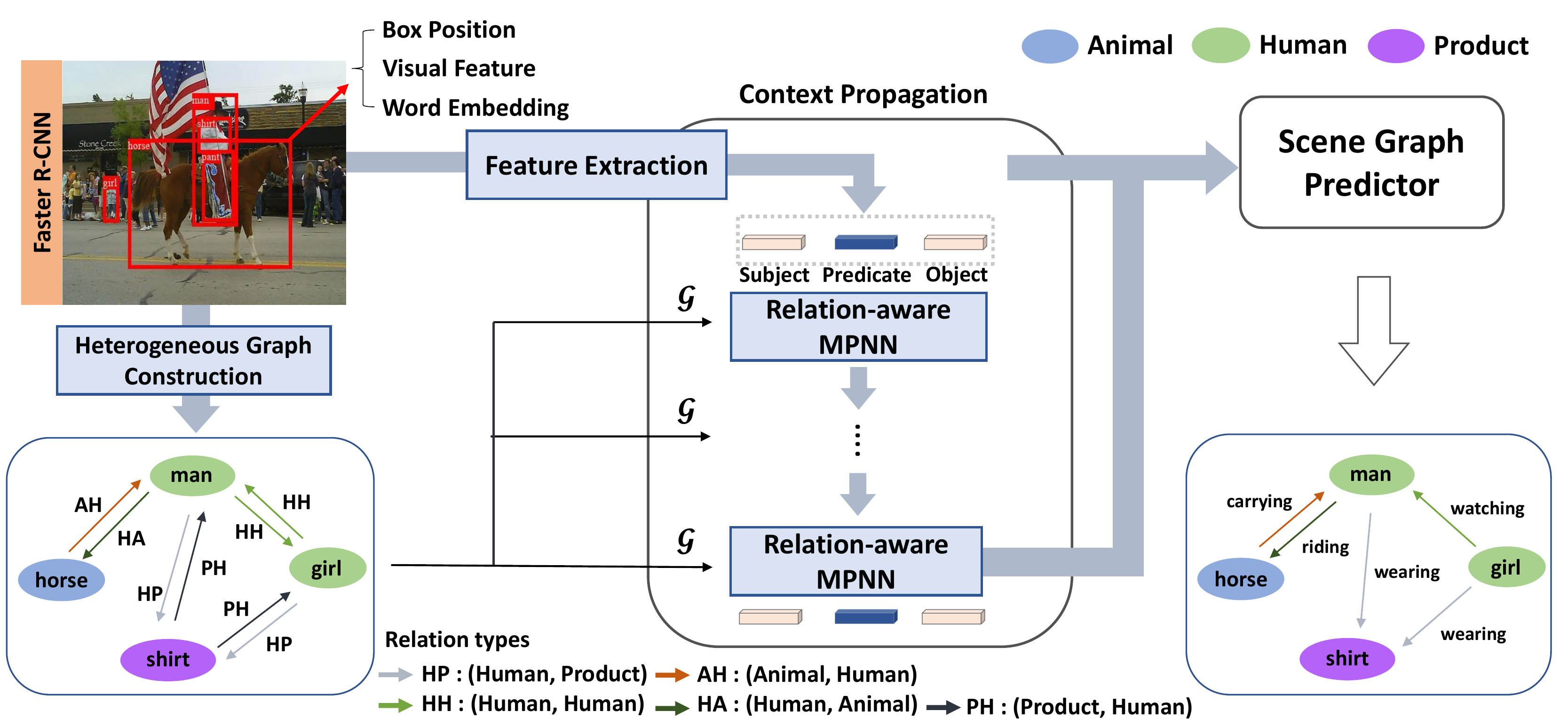}
    \end{center}
    \caption{Given an image, a heterogeneous graph is constructed based on the objects detected by an object detector (i.e., Faster R-CNN) from which feature vectors for objects and predicates are extracted. RMP propagates relation-aware messages to the representations of objects and predicates. Finally, the scene graph predictor generates a heterogeneous scene graph.
    }
    \label{fig:overall}
    
\end{figure*}

\section{Related Work}
\noindent\textbf{Scene Graph Generation.} Traditional SGG approaches mainly focus on designing advanced neural network architectures to fuse the contextual features in images. They \cite{chen2019knowledge,neural_motif} pass object features of region proposals obtained from an object detector to RNNs aiming at capturing the contextual cues. However, contextualized representations produced by RNNs are dependent on the ordering of the object sequence, which is generated in a heuristic manner without accounting for meaningful contexts.
Most recently, MPNN-based methods have shown to be effective for SGG tasks. With an effective architecture that aggregates the neighboring features, MPNNs embed a node-centric representation by combining the contextual elements of neighbors. Several variants of MPNNs have emerged for SGG frameworks as it is essential to design MPNNs with certain characteristics.
GPS-Net \cite{gps_net} proposes the direction-aware MPNN (DMP) to propagate different messages in two directions, and BGNN \cite{bgnn} presents confidence-aware adaptive MPNN (AMP) to filter out unnecessary contextual features because the graph constructed from an off-the-shelf object detector contains noise. 
However, the assumption of the existing methods that object and relation types are all equal restricts their performance as the semantics of the relations cannot be captured. In this work, we introduce a new scene graph structure that is based on a heterogeneous graph, and focus on designing a novel MPNN layer, i.e., relation-aware MPNN (RMP), that captures the semantics of predicate types.

\smallskip
\noindent\textbf{Heterogeneous Graph Neural Network.} A heterogeneous graph is a powerful tool that can embrace rich semantics and structural information in real-world data. A heterogeneous graph consists of different types of entities (i.e., nodes) and their relations (i.e., edges), 
which facilitates the modeling of complex semantics of machine learning models. In the graph mining community, a plethora of studies for heterogeneous graph neural networks (HGNN) have been conducted to extract rich information under the heterogeneity by using the \textit{meta-path}, which is a semantically meaningful path defined by relation types~\cite{HAN,Pathsim,metapath2vec,hgt,hetgnn,park2019task}.
As a scene graph represents multiple objects and the relations between them, we argue that it can also be considered as a heterogeneous graph. 
However, it is non-trivial to apply existing HGNNs developed in the graph mining community for SGG tasks. More precisely, although most existing HGNNs require domain knowledge to generate meta-paths, there are no obvious rules for defining meta-paths in scene graphs.
To make the matter worse, since the graph constructed from an off-the-shelf object detector is inherently noisy, it is challenging to apply existing HGNNs that are developed for clean graphs.
Hence, in this work, we propose a meta-path-free HGNN that considers the noisy nature of scene graphs.
Although there have been comprehensive studies of MPNN layers for SGG, they regard a scene graph as a homogeneous graph, and overlook the node types and predicate types.
To the best of our knowledge, this is the first work to consider a scene graph in the SGG task as a heterogeneous graph.

\smallskip
\noindent\textbf{Long-tail Visual Recognition.} 
Recent SGG methods mainly focus on relieving the long-tail problem of the predicate class distribution for constructing informative scene graphs. One of the prominent approaches to alleviate the long-tail problem is to employ a cost-sensitive loss for SGG. More precisely, some methods introduce novel reweighted losses that leverage semantic constraints of scene graphs~\cite{knyazev2020graph,gps_net}. 
Moreover, BGNN \cite{bgnn} proposes a powerful data re-sampling strategy, called bi-level sampling, which combines both image-level and instance-level re-sampling strategies, aiming to make the entire training predicate class distribution towards a uniform distribution. 
Recently, TDE \cite{unbiased} utilizes causal inference in the prediction stage to alleviate the effect of the long-tail predicate class distribution of the dataset. 
On the other hand, as illustrated in Figure~\ref{fig:fig_1}(c), our proposed method naturally alleviates the long-tail problem by considering a scene graph as a heterogeneous graph.

\section{Problem Definition}
In this section, we describe notations used throughout the paper, and introduce our formulation of the SGG task. Let $\mathcal{G}= <\mathcal{V}, \mathcal{E}, \mathcal{T}_\mathcal{V}, \mathcal{T}_\mathcal{E}>$  be a heterogeneous graph, where $\mathcal{V}$ is the set of objects in an image, $\mathcal{E}$ is the set of relations between objects, $\mathcal{T}_\mathcal{V}$ is the set of object types, and $\mathcal{T}_\mathcal{E}$ is the set of relation types. Objects $u, v \in \mathcal{V}$ and the relation $e_{u\rightarrow v} \in \mathcal{E}$ between $u$ and $v$ have feature vectors $x_u$, $x_v$ and $x_{u\rightarrow v}$, respectively.
Moreover, we denote $\mathcal{Y}=<\mathcal{Y}_o, \mathcal{Y}_r>$ as the set of $|\mathcal{Y}_o|$ object classes and $|\mathcal{Y}_r|$ relation (i.e., predicate) classes.

Our goal is to find a function that maps an image $I$ to a scene graph by maximizing the probability $P(
\mathcal{G}, \mathcal{Y}|\mathcal{I})$. Specifically, we aim to estimate the graph structural information $\mathcal{G}$, the object classes (i.e., $\mathcal{Y}_o$) and the relation classes (i.e., $\mathcal{Y}_r$) given an image $\mathcal{I}$.
In this work, we introduce a novel framework for generating scene graphs, called \textsf{\textbf{Het}}erogeneous \textsf{\textbf{S}}cene \textsf{\textbf{G}}raph \textsf{\textbf{G}}eneration (\proposed), which generates a scene graph with typed objects and relations.
$P(\mathcal{G}, \mathcal{Y}|\mathcal{I})$ can be factorized as follows: $P(\mathcal{G}, \mathcal{Y}|\mathcal{I}) =$ $P(\mathcal{G}|\mathcal{I})$$P(\mathcal{Y}_o|\mathcal{G}, \mathcal{I})$$P(\mathcal{Y}_r|\mathcal{Y}_o, \mathcal{G}, \mathcal{I})$,
where $P(\mathcal{G}|\mathcal{I})$ is a heterogeneous graph construction module,  $P(\mathcal{Y}_o|\mathcal{G}, \mathcal{I})$ is an object classifier, and  $P(\mathcal{Y}_r|\mathcal{Y}_o, \mathcal{G}, \mathcal{I})$ is a predicate classifier.
Note that our formulation is different from that of recent SGG methods~\cite{VCTree,bgnn,unbiased,chen2019knowledge,gps_net,VCTree} in terms of predicting and utilizing the heterogeneous information $\mathcal{T}_\mathcal{V}$ and $\mathcal{T}_\mathcal{E}$ of $\mathcal{G}$. Thus, our framework is a generalized version of existing homogeneous SGG methods (i.e.,~\proposed~degenerates to existing homogeneous SGG methods if $|\mathcal{T}_\mathcal{V}|=|\mathcal{T}_\mathcal{E}|=1$.). The overall architecture of~\proposed~is described in Figure~\ref{fig:overall}.

\section{Methodology}
\subsection{Heterogeneous Graph Construction}
\noindent\textbf{Initial Graph Construction.}
We begin by constructing an initial graph based on the objects detected by an off-the-shelf object detector (e.g., Faster R-CNN \cite{faster_rcnn})\footnote{Although the model performance can be improved by using more advanced object detectors, we rely on Faster R-CNN to clearly validate the benefit of our framework.}. The object proposals generated by the object detector are defined as nodes, and node pairs are connected by edges (i.e., the initial graph is a fully-connected graph). Then, the feature vector for object $u\in\mathcal{V}$ (i.e., $x_u$) is obtained by feed-forwarding the concatenation of the bounding box positions of object $u$, visual features of object $u$, and the word embedding (i.e., Glove \cite{glove}) of the class name of object $u$. 
Moreover, the feature vector of the relation between object $u$ and $v$ (i.e., $x_{u\rightarrow v}$) is extracted from the bounding box positions of $e_{u\rightarrow v}$, and the visual features of the union box of the object pairs $(u,v)$.

\begin{figure*}[t]
    \begin{center}
        \includegraphics[width=.65\linewidth]{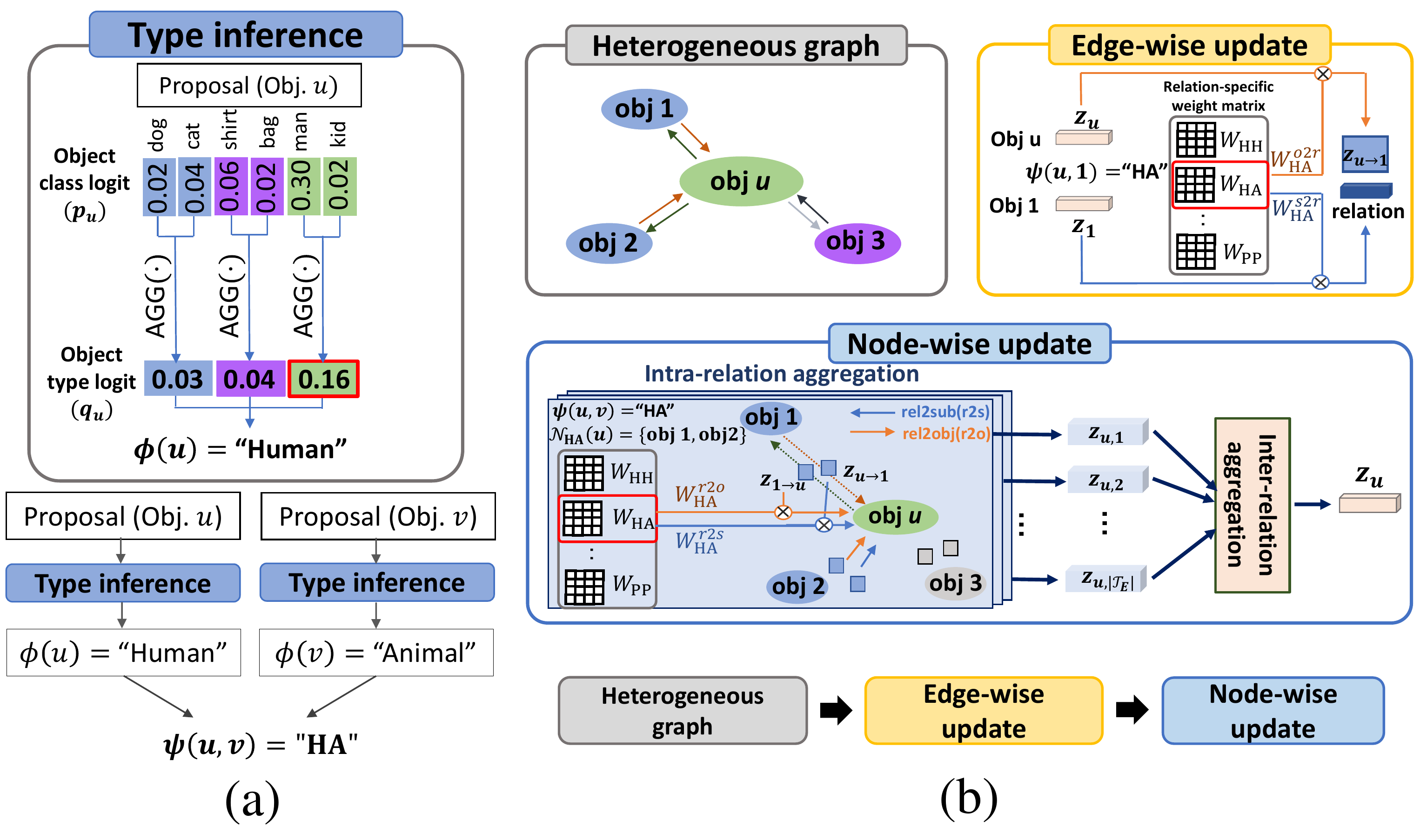}
    \end{center}
    \vspace{-2ex}
    \caption{
    \textbf{(a) Type assignment.} 
    From the class logits of an object proposal, object types and relation types are determined. Type assignment module computes its probability for each object type with pre-defined mapping. The Cartesian product of object types are defined as relation type.
    \textbf{2) RMP.} 
    For edge-wise update,  relation type-specific weight matrices are used to generate two-way messages, which are then used to obtain the relation representation.
    For the node-wise update, the intra-relation aggregation produces the relation-specific messages using the predicate features. Then, the inter-relation aggregation gathers all the relation information to make the relation-aware object representation. 
    }
    \label{fig:fig3}
    \vspace{-1ex}
\end{figure*}

\smallskip
\noindent\textbf{Type Inference.}
To convert the initial graph into a heterogeneous graph, we assign types to the objects and the relations by utilizing the class logits obtained by the object detector.
More precisely, the proposal for object $u$ contains the class logit $p_u \in\mathbb{R}^{|\mathcal{Y}_o|}$ obtained from the object detector, where each element of $p_u$ denotes the logit value for a certain object class. We compute the object type logit vector $q_u \in \mathbb{R}^{|\mathcal{T}_\mathcal{V}|}$ with a pre-defined function $\phi$ that maps an object class to an object type, i.e., $\phi:\mathcal{Y}_o \rightarrow \mathcal{T}_\mathcal{V}$. We then infer the type for object $u$ with a simple aggregation function, such as sum, mean, and max as in Figure~\ref{fig:fig3}(a). For example, consider two detected objects $u$ and $v$, which belong to \textsf{man} and \textsf{dog} object classes, respectively, according to the class logits $p_u$ and $p_v$. Then, the object types are obtained as: $\phi(u)=\text{``Human'' (H)}$ and $\phi(v)=\text{``Animal'' (A)}$. Note that we employ \textsf{Average($\cdot$)} as the aggregation function, but we empirically observed that \textsf{Sum($\cdot$)} performs similarly. Lastly, the type of the relation (i.e., predicate type) between objects $u$ and $v$ (i.e., $e_{u\rightarrow v}$) is automatically determined by the associated object types with a function $\psi: \mathcal{Y}_o \times \mathcal{Y}_o \rightarrow \mathcal{T}_\mathcal{E}$, which is $\psi(u, v)=\text{``HA''}$ in our example. In this work, we mainly consider three object types, i.e., ``Human (H)'', ``Animal (A)'', and ``Product (P)'', which consequently produce nine predicate types, i.e., ``HH'', ``HA'', ``HP'', ``AH'', ``AA'', ``AP'', ``PA'', ``PH'', and ``PP.'' 

It is important to note that since the heterogeneous graph on which~\proposed~is applied is constructed solely based on the above type inference process, accurately inferring the types is crucial for the performance of~\proposed. In Table~\ref{table:analysis_type} of Section~\ref{sec:rmp}, we empirically show that through the above type assignment process based on Faster R-CNN as the object detector, we achieve around 95\% accuracy of the object type inference, and we demonstrate further improvements of~\proposed~when ground-truth object types are used, i.e., when the object type inference accuracy is 100\%. Furthermore, we show that adding another predicate type, i.e., ``Landform (L)'', further improves the model performance provided that the type inference is accurate.

\subsection{Relation-aware Contextual Representation Learning}
Now that we have constructed a heterogeneous graph, we need to learn the representations of objects and their relations. 
To this end, we propose a novel message passing layer, called \textbf{R}elation-aware \textbf{M}essage \textbf{P}assing neural network (RMP), to capture relation-aware context. 
The main idea is to treat each relation differently according to its type.
A na\"ive approach is to assign a type-specific projection matrix $W_t$ to each relation type $t\in\mathcal{T}_\mathcal{E}$.
However, using a distinct parameter for each type is not only computationally inefficient as the model complexity grows proportional to the number of relation types $\mathcal{T}_\mathcal{E}$, but also is prone to overfitting.
Hence, we compose an 
efficient relation type-specific projection matrix $W_t\in\mathbb{R}^{d\times d}$ as a linear combination of shared basis matrices~\cite{rgcn}: $W_t = \sum_{i=1}^b a_{ti}B_i$,
where $B_i\in\mathbb{R}^{d\times d}$ is a trainable matrix for basis $i$, $b$ is the number of bases, and $a_{ti}\in\mathbb{R}$ is a trainable coefficient for relation type $t$ and basis $i$, which captures the relation type information.
A large $a_{ti}$ implies that $B_i$ makes a large contribution when we compose the relation type-specific projection matrix $W_t$.
Based on the reformulation of relation type-specific projection matrix,
we can greatly reduce the number of parameters (i.e., from $\mathcal{O}(d^2|\mathcal{T}_\mathcal{E}|)$ to $\mathcal{O}(d^2b + |\mathcal{T}_\mathcal{E}|b)$, where $b\ll|\mathcal{T}_\mathcal{E}|$), which not only facilitates efficient training, but also alleviates the overfitting issue.

The goal of RMP is to capture the relation-aware context and update the representation of objects and relations using the context. In this regard, RMP consists of the following two steps: \textbf{1) Edge-wise update for relations}, and \textbf{2) Node-wise update for objects}. 
In a nutshell, in the  edge-wise  update step, RMP generates relation-specific messages between objects to refine the relation representations.
In the node-wise update step, RMP aggregates messages from neighboring relations according to the relation types (i.e., intra-relation aggregation), and then aggregates all the relation type-specific object representations to obtain the final object representations (i.e., inter-relation aggregation). 
Note that \textit{RMP is model-agnostic in that it can be adopted to any MPNN-based SGG methods}, such as Graph R-CNN~\cite{graph_rcnn} and BGNN~\cite{bgnn}. Due to the space limitation, we only explain RMP adopted to Graph R-CNN (i.e.,~\proposed) in this paper. Refer to Appendix~\ref{appendix:bgnn} for details on RMP adopted to BGNN (i.e., $\text{\proposed}_{\texttt{++}}$)

\smallskip
\noindent \textbf{Step 1) Edge-wise update for relations.} 
In this step, given two objects $u$ and $v$, and the relation between the objects, i.e., $e_{u\rightarrow v}$, the objects propagate contextual information to the relation.
More precisely, two directional messages are generated: one is from object $u$ to relation $e_{u\rightarrow v}$, and the other is from object $v$ to relation $e_{u\rightarrow v}$.
For an edge $e_{u\rightarrow v}$ whose direction is from object $u$ to object $v$, the relation representation is computed as follows:

\begin{equation}
\small
    z_{u\rightarrow v}^{(l+1)} = z_{u\rightarrow v}^{(l)} + \sigma ( \alpha(u,v) W_{\psi(u,v)}^{\textsf{s2r}} z_u^{(l)} + (1-\alpha(u,v)) W_{\psi(u,v)}^{\textsf{o2r}} z_v^{(l)})
\end{equation}
where $z_{u\rightarrow v}^{(l+1)}\in\mathbb{R}^d$ is the relation representation of $e_{u\rightarrow v}$ at the $(l+1)$-th layer, $W_{\psi(u,v)}^{\textsf{s2r}}\in\mathbb{R}^{d\times d}$ and $W_{\psi(u,v)}^{\textsf{o2r}}\in\mathbb{R}^{d\times d}$ are weight matrices of the relation type $\psi(u,v)$ for the two-way messages (i.e., \textsf{subject}-to-\textsf{relation} and \textsf{object}-to-\textsf{relation} messages given a triplet $\langle$\textsf{subject}, \textsf{predicate}, \textsf{object}$\rangle$), and $\sigma$ is a non-linear activation function. The initial representations for objects and relations are feature vectors of objects and relations (i.e., $z_{u\rightarrow v}^{(0)}=x_{u\rightarrow v}$ and $z_u^{(0)}=x_u$). 
Moreover, $\alpha(u,v)$ determines the importance of the messages propagated from \textsf{subject} (i.e., $u$) and \textsf{object} (i.e., $v$), and it is formulated as follows: 
$\alpha(u,v)=\frac{\text{exp}(w^Tz_u^{(l)})}{\text{exp}(w^Tz_u^{(l)})+\text{exp}(w^Tz_v^{(l)})}$,
where $w\in\mathbb{R}^d$ is an attention vector.
A large $\alpha(u,v)$ implies that object $u$ is more important than object $v$ for generating the representation of the relation $e_{u\rightarrow v}$.

\begin{table*}[t]
\centering
\small
\begin{tabular}{c|cc|cc|cc}
\hline
\multirow{2}{*}{\textbf{Models}} & \multicolumn{2}{c|}{\textbf{PredCls}}  & \multicolumn{2}{c|}{\textbf{SGCls}}    & \multicolumn{2}{c}{\textbf{SGGen}}     \\ \cline{2-7} 
                                 & \textbf{mR@50/100} & \textbf{R@50/100} & \textbf{mR@50/100} & \textbf{R@50/100} & \textbf{mR@50/100} & \textbf{R@50/100} \\ \hline
RelDN \cite{RelDN}                            & 15.8/17.2          & 64.8/66.7         & 9.3/9.6            & 38.1/39.3         & 6.0/7.3            & 31.4/35.9         \\
Motifs \cite{neural_motif}                           & 14.6/15.8          & 66.0/67.9         & 8.0/8.5            & 39.1/39.9         & 5.5/6.8            & 32.1/36.9         \\
VCTree \cite{VCTree}                          & 15.4/16.6          & 65.5/67.4         & 7.4/7.9            & 38.9/39.8         & 6.6/7.7            & 31.8/36.1         \\
G-RCNN \cite{graph_rcnn}                           & 16.4/17.2          & 65.4/67.2         & 9.0/9.5            & 37.0/38.5         & 5.8/6.6            & 29.7/32.8         \\
MSDN \cite{msdn}                             & 15.9/17.5          & 64.6/66.6         & 9.3/9.7            & 38.4/39.8         & 6.1/7.2            & 31.9/36.6         \\
Unbiased \cite{unbiased}                         & 25.4/28.7          & 47.2/51.6         & 12.2/14.0          & 25.4/27.9         & 9.3/11.1           & 19.4/23.2         \\
GPS-Net \cite{gps_net}                          & 15.2/16.6          & 65.2/67.1         & 8.5/9.1            & 37.8/39.2         & 6.7/8.6            & 31.1/35.9         \\
$\text{GPS-Net}^{\ddagger}$ \cite{gps_net}                         &    29.2/31.4      &     55.2/57.6     & 15.9/16.9            & 36.4/37.5         & 8.1/9.6            & 28.4/33.4         \\

NICE-Motif\cite{NICE} & 29.9/32.3 & 55.1/57.2 & 16.6/17.9  & 33.1/34.0 & \textbf{12.2}/\textbf{14.4} & 27.8/31.8 \\
PPDL\cite{PPDL} & 32.2/33.3 & 47.2/47.6 & 17.5/18.2 & 28.4/29.3 & 11.4/13.5 & 21.2/23.9 \\

$\text{BGNN}^{\ddagger}$ \cite{bgnn}                            & 30.4/32.9          & 59.2/61.3         & 14.3/16.5          & 37.4/38.5         & 10.7/12.6          & 31.0/35.8         \\ 

$\text{BGNN}^{\ast \ddagger}$ \cite{bgnn}                            & 29.2/31.7          & 57.8/60.0         & 14.6/16.0          & 36.9/38.1         & 10.9/13.1          & 30.2/34.9         \\ \hline
$\proposed^{\ddagger}$                           & 31.6/33.5          & 57.8/59.1         & \textbf{17.2/18.7}       & 37.6/38.7         & \textbf{12.2/14.4}          & 30.0/34.6         \\ 
$\text{\proposed}^{\ddagger}_\texttt{++}$                           & \textbf{32.3/34.5}          & 57.1/59.4         & 15.8/17.7          & 37.6/38.5         & 11.5/13.5          & 30.2/34.5         \\
\hline
\textbf{Improv.(\%)}  & \textbf{10.6/8.8} & 0.0/-1.0 &   \textbf{17.8/16.9} & 1.9/1.6                & \textbf{11.9/9.9} & 0.0/-0.8                 
\end{tabular}
\vspace{-1ex}
\caption{Results on Visual Genome \cite{visualgenome}. \textbf{Improv.} denotes improvements of~\proposed~compared with BGNN$^{\ast \ddagger}$. 
$\ddagger$ denotes bi-level sampling~\cite{bgnn} is applied, and $\ast$ denotes results reproduced with authors' code.
}
\vspace{-3ex}
\label{table:main_table}
\end{table*}

\smallskip
\noindent \textbf{Step 2) Node-wise update for objects.} 
The node update is performed based on the relation representation $z_{u\rightarrow v}^{(l+1)}$ obtained in the previous step. The main idea is to aggregate messages from neighboring objects that share the same relation type. More precisely, given an object $u$ and its neighboring objects with the relation type $t$ (i.e., $\mathcal{N}_t(u)$), the representation of the object $u$ regarding the relation type $t$ at $(l+1)$-th layer (i.e., $z_{u, t}^{(l+1)}\in\mathbb{R}^d$) is computed as follows: 

\begin{equation}
\small
    z_{u, t}^{(l+1)} = \sum_{v\in \mathcal{N}_t(u)}  \alpha_{\textsf{r2s}}(v, t) W_{\psi(u,v)}^{\textsf{r2s}} z_{u\rightarrow v}^{(l+1)} + \alpha_{\textsf{r2o}}(v, t) W_{\psi(u,v)}^{\textsf{r2o}} z_{v\rightarrow u}^{(l+1)}
\end{equation}
where $W_{\psi(u,v)}^{\textsf{r2s}}\in\mathbb{R}^{d\times d}$ and $W_{\psi(u,v)}^{\textsf{r2o}}\in\mathbb{R}^{d\times d}$ are weight matrices for the \textsf{relation}-to-\textsf{subject} and \textsf{relation}-to-\textsf{object} messages, respectively.
The first term produces a message in the perspective of $u$ being the \textsf{subject} in a triplet, and the second term produces a message in the perspective of $u$ being the \textsf{object} in a triplet. We denote the above process as \textit{intra-relation aggregation}. 
Moreover, $\alpha_{\textsf{r2s}}(v,t)$ and $\alpha_{\textsf{r2o}}(v,t)$ denote the importance of object $v$ among the set of neighbors of object $u$ with relation type $t$ (i.e., $\mathcal{N}_t(u)$). They are defined as follows:
$\alpha_{\textsf{r2s}}(v,t) = \frac{\text{exp}(w_{\textsf{r2s,t}}^{T} z_{u\rightarrow v}^{(l+1)})}{\sum_{q\in\mathcal{N}_t(u)}\text{exp}(w_{\textsf{r2s,t}}^{T} z_{u\rightarrow q}^{(l+1)})},$
$\alpha_{\textsf{r2o}}(v,t) = \frac{\text{exp}(w_{\textsf{r2o,t}}^{T} z_{v\rightarrow u}^{(l+1)})}{\sum_{q\in\mathcal{N}_t(u)}\text{exp}(w_{\textsf{r2o,t}}^{T} z_{q\rightarrow u}^{(l+1)})}$
, where $w_{\textsf{r2s,t}}$ and $w_\textsf{r2o,t} \in \mathbb{R}^d$ are attention vectors for \textsf{relation-to-subject} updates and \textsf{relation-to-object} updates considering the relation type $t$, respectively.
A large $\alpha_{\textsf{r2s}}(v,t)$ and $\alpha_{\textsf{r2o}}(v,t)$ imply that under the relation type $t$, the relation from $u$ to $v$ (i.e., $e_{u\rightarrow v}$) and the relation from $v$ to $u$ (i.e., $e_{v\rightarrow u}$) are crucial for object $u$, respectively.
Finally, we aggregate all the relation type-specific object representations to obtain the final representation of object $u$ (i.e., $z_{u}^{(l+1)}\in\mathbb{R}^d$):
$z_u^{(l+1)} = z_u^{(l)} + \frac{1}{|\mathcal{T}_\mathcal{E}|}\sum_{t=1}^{|\mathcal{T}_\mathcal{E}|} \sigma(z_{u,t}^{(l+1)}).$
We denote the above process as \textit{inter-relation aggregation}.
In summary,~\proposed~generates relation-specific context through intra- and inter-relation aggregations. Moreover, high-order interactions between objects can be captured by stacking multiple RMP layers.

\begin{figure}[t]
    \centering
    \includegraphics[width=0.95\columnwidth]{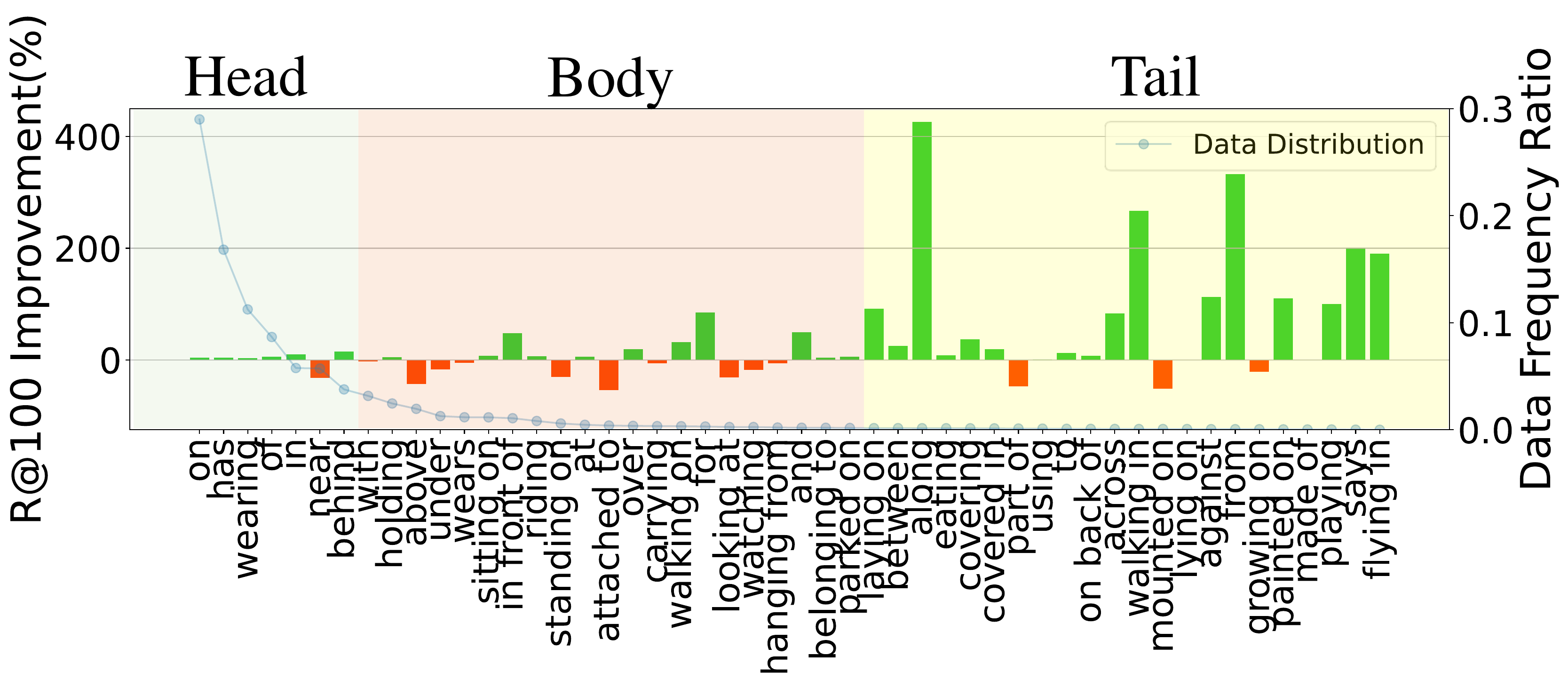}
    \caption{R@100 improvement per class of~$\text{{\proposed}}^{\ddagger}$~over $\text{BGNN}^{\ast \ddagger}$ in SGCls task.
    }
    \label{fig:per_class_sgcls}
    \vspace{-1.5ex}
\end{figure}

\subsection{Scene Graph Predictor and Model Training}
\noindent\textbf{Scene Graph Predictor.}
After obtaining the representations for objects and relations of an image, we predict their classes to generate the scene graph of the image.
First, given the representation of object $u$ (i.e., $z_u$), we employ a simple linear classifier to obtain the object class probability as follows: $p_u = \text{softmax}(W^{\text{obj}} z_u) \in \mathbb{R}^{|\mathcal{Y}_o|}$,
where $W^{\text{obj}}\in\mathbb{R}^{|\mathcal{Y}_o|\times d}$ is the weight matrix for the linear classifier.
Next, given the representation of the relation between objects $u$ and $v$ (i.e., $z_{u\rightarrow v}$), we employ a simple linear classifier along with an added bias term regarding the class frequency prior (i.e., $\hat{p}_{u\rightarrow v}$)~\cite{gps_net,bgnn,neural_motif}. The relation class probability is computed as follows: $p_{u\rightarrow v} = \text{softmax}(W^{\text{rel}}z_{u \rightarrow v} + \log{\hat{p}_{u\rightarrow v}}) \in \mathbb{R}^{|\mathcal{Y}_r|}$,
where $W^{\text{rel}}\in\mathbb{R}^{|\mathcal{Y}_r|\times d}$ is the weight matrix for the linear classifier, and $\hat{p}_{{u\rightarrow v}}\in\mathbb{R}^{|\mathcal{Y}_r|}$ denotes the frequency distribution of predicates given two objects $u$ and $v$, which is pre-computed from the training data.

\smallskip
\noindent\textbf{Model Training.}
Using the object class probability (i.e., $p_{u}$) and relation class probability (i.e., $p_{u\rightarrow v}$),~\proposed~is trained by minimizing the conventional cross-entropy losses for objects (i.e., $\mathcal{L}_{\text{obj}}$) and relations (i.e., $\mathcal{L}_{\text{rel}}$).
The final objective function is defined as follows: $\mathcal{L}_{\text{final}} = \mathcal{L}_{\text{obj}} + \mathcal{L}_{\text{rel}}$.

\section{Experiment}
We evaluate~\proposed~compared with state-of-the-arts methods on commonly used benchmark datasets, i.e., Visual Genome (VG) \cite{visualgenome} (\textbf{Section~\ref{sec:vg}}), and Open Images (OI) V6 \cite{openimage} (\textbf{Section~\ref{sec:oi_v6}}). More details on each dataset are described in Appendix \ref{appendix:details}.

\smallskip
\noindent\textbf{Evaluation Metric.}
Due to the long-tail problem in SGG tasks, existing methods perform poorly on less frequently appearing predicates. Therefore, following the evaluation protocol of recent SGG methods~\cite{unbiased}, we evaluate SGG models on \textbf{mean Recall@K (mR@K)} in addition to the conventional measure \textbf{Recall@K (R@K)}~\cite{msdn}.
For Open Images dataset, we follow the evaluation protocols of previous works ~\cite{openimage,gps_net}, and we additionally report weighted mean AP of relationships ($\text{wmAP}_{\text{rel}}$), weighted mean AP of phrase ($\text{wmAP}_{\text{phr}}$), and the weighted metric score ($\text{score}_{\text{wtd}}$), which is calculated as: $0.2 \times \text{R@50} + 0.4 \times \text{wmAP}_{\text{rel}}+0.4\times \text{wmAP}_{\text{phr}}$.

\begin{figure}
    \centering
    \includegraphics[width=.93\columnwidth]{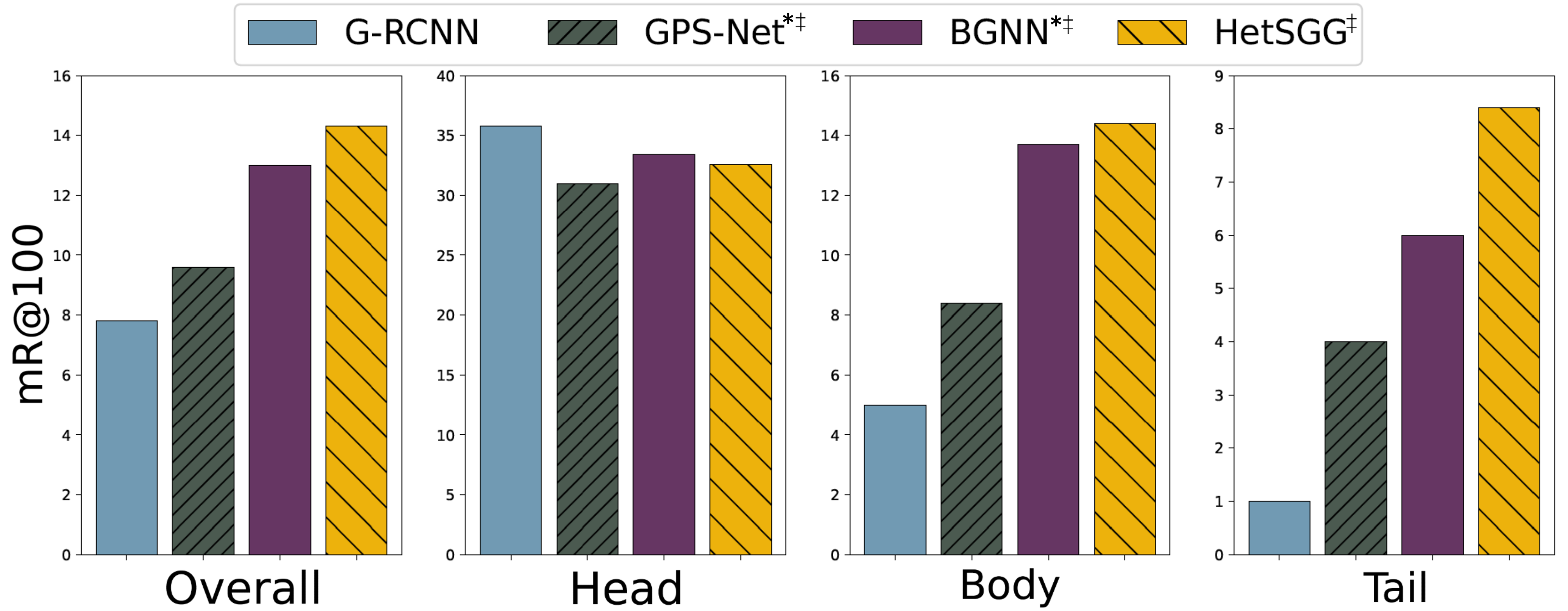}
    \caption{Results on the overall, head, body, and tail predicate classes in SGGen task.
    }
    \label{fig:per_class_sggen_hbt}
    \vspace{-1.2ex}
\end{figure}

\smallskip
\noindent\textbf{Evaluation Protocol.} We evaluate on three conventional SGG tasks~\cite{IMP}:
(1) \textbf{Predicate Classification (PredCls)},
(2) \textbf{Scene Graph Classification (SGCls)}, and
(3) \textbf{Scene Graph Generation (SGGen)}.
Note that for {SGGen}, an object is considered to be correctly detected when its IoU (Intersection over Union) with the ground truth bounding box is greater than 0.5.

\smallskip

\noindent\textbf{Implementation Details.} 
For fair comparisons, we adopt ResNeXt-101-FPN \cite{resnet_101} and Faster R-CNN \cite{faster_rcnn} as the object detector, whose parameters are frozen while training the SGG model.
For {SGGen} task, we select the top 80 object proposals sorted by object scores, and use per-class non-maximal suppression (NMS)~\cite{neural_motif} at IoU 0.5.
To obtain the relation proposals, we localize the union box of the bounding boxes of two objects, and obtain the ROI features using pre-trained Faster R-CNN. 
For RMP of~\proposed, we set the number of bases (i.e., b) to 8 in VG and 4 in OI, and use four MPNN layers (i.e., l = 4). 
Since~\proposed~framework is model-agnostic, we evaluate two versions of~\proposed: i) RMP adopted to Graph R-CNN \cite{graph_rcnn} (i.e., HetSGG), and ii) RMP adopted to BGNN \cite{bgnn} (i.e., $\proposed_{\texttt{++}}$). For more details of hyper-parameter setting, refer to Appendix \ref{appendix:hyper_config}.


\subsection{Visual Genome}
\label{sec:vg}

\subsubsection{Dataset Details.} We follow the same pre-processing strategy that has been widely used for evaluations of SGG \cite{IMP}. Specifically, the most frequently appearing 150 object classes and 50 predicate classes are used for evaluation.
After preprocessing, each image contains 11.6 objects and 6.2 predicates on average.
A total of 108k images are split into training set (70\%) and test set (30\%). 

\subsubsection{Comparisons with State-of-the-Art Methods. }
Table~\ref{table:main_table}~shows the results on various SGG tasks in terms of mR@50/100 and R@50/100. We have the following observations:
\textbf{1) }{\proposed~and~$\proposed_\texttt{++}$} generally outperform all baseline models in various tasks on both metrics. More precisely,{~\proposed~and~$\proposed_{\texttt{++}}$} greatly improve mR@50/100, while performing competitively on R@50/100. This verifies that~\proposed~effectively relieves the long-tail problem of the predicate class distribution by considering scene graphs as heterogeneous graphs.
\textbf{2) }It is important to note that for fair comparisons among various MPNN architectures (i.e., DMP of GPS-Net~\cite{gps_net}, AMP of BGNN~\cite{bgnn}, and RMP of~\proposed), we also compare with the version of GPS-Net to which bi-level sampling~\cite{bgnn} is applied (i.e., $\text{GPS-Net}^{\ddagger}$).
We observe that both {\proposed~and~$\proposed_{\texttt{++}}$} outperform $\text{GPS-Net}^{\ddagger}$ and $\text{BGNN}^{\ast \ddagger}$, which demonstrates the superiority of RMP of~\proposed.
In summary, these results imply that using the heterogeneous information inherent in scene graphs has more powerful predictive performance compared with the methods that utilize MPNNs based on homogeneous scene graphs.

Moreover, Figure~\ref{fig:per_class_sgcls} shows improvements of~\proposed~over BGNN per predicate class.
The order of the predicates (i.e., $x$-axis) is sorted by the frequency of predicates in the training data. We observe that~\proposed~generally achieves improvements on all head, body and tail predicate classes, while particularly showing great improvements on tail predicate classes. Similar results are shown in Figure~\ref{fig:per_class_sggen_hbt}. It is important to note that there is a clear trade-off between the head and tail performance as shown in existing studies~\cite{chen2022resistance,Desai_2021_ICCV,bgnn}.
However, since the predicates in head classes (e.g., on, has, of, in, etc) are less informative for generating meaningful scene graphs, it is important to achieve a high performance on tail classes. In this regard,~\proposed~is superior to other baselines in that \textit{it outperforms baselines in terms of tail performance,  while also maintaining a competitive head performance}.


                             

\begin{table}[h]
\centering
\small
\begin{tabular}{c|p{1.35cm}|p{1.35cm}p{1.35cm}|c}
\hline
\multirow{3}{*}{\textbf{\begin{tabular}[x]{@{}c@{}} Object\\Types\end{tabular}}}  &   & \multicolumn{2}{c|}{\textbf{SGCls}} &   \text{\textbf{Type}}  \\ \cline{3-4}

 &   \textbf{Model} & \multirow{2}{*}{\text{\textbf{mR@50/100}}} & \multirow{2}{*}{\text{\textbf{R@50/100}}} & \text{\textbf{Inf.}}   \\ 
& & & & \text{\textbf{Acc.}}\text{(\%)} \\ \hline
                             
\multirow{2}{*}{P,H,A}           &  $\text{\proposed} ^{\ddagger}$                     &      17.2 / 18.7       &       37.6 / 38.7   &  95.3    \\
     & $\text{\proposed}_{\text{GT}}^{\ddagger}$                     &     17.4 / 19.1          &      38.0 / 39.0  & 100        \\ \hline
\multirow{2}{*}{P,H,A,L} & $\text{\proposed}^{\ddagger}$                       &       15.9 / 18.2             &      37.5 / 38.4   & 90.9       \\
 & $\text{\proposed}_{\text{GT}}^{\ddagger}$                    &       18.2 / 19.4          &   39.4 / 40.5   & 100   \\    \hline \hline
\end{tabular}

\caption{Analysis on the object types and the accuracy of the type inference module when different object types are defined: P-Product, H-Human, A-Animal, L-Landform.}
\vspace{-2ex}
\label{table:analysis_type}
\end{table}

\subsubsection{Analysis on Object Types. }
\label{sec:rmp}
\noindent In this section, we conduct the following two experiments, and show results in Table~\ref{table:analysis_type}:
i) To investigate the importance of accurate object type inference on the performance of~\proposed, we train~\proposed~using the ground-truth object types (i.e.,${\proposed}_{\text{GT}}$), and ii) To investigate the benefit of considering the object types in a more fine-grained manner, we add another object type, i.e., ``Landform\footnote{We sample ``Landform'' predicates from ``Product'' predicates, while ``Human'' and ``Animal'' predicates remain unchanged.}.''
We have the following observations:
\textbf{1)} ${\proposed}_{\text{GT}}$ consistently outperforms~\proposed. This implies that accurately inferring the object types is crucial, and that an advanced object detector can further improve~\proposed.
\textbf{2)} Comparing the performance of ${\proposed}_{\text{GT}}$, we observe that considering four object types is superior to considering three object types. This implies that it is beneficial to consider the object types in a more fine-grained manner. 
\textbf{3)} However, comparing the performance of~\proposed s, we observe that when the type inference accuracy is not high enough, the performance degrades when considering more types. This again demonstrates the importance of accurately inferring the object types. In the same context, we analyze how the performance is affected by aggregation function, e.g., sum, max, which outputs the different type inference accuracy in Appendix \ref{appendix:hyper}.

\begin{table}[t]
    \small
        \centering
        
        \begin{tabular}{cc|cc}
            \hline
            \multicolumn{2}{c|}{Component} & \multicolumn{2}{c}{Metric} \\
            \hline
            Edge & Node & mR@100 & R@100 \\
            \hline
             \ding{55} & \ding{55} & 15.9 & 38.6 \\
             \ding{51} & \ding{55}  & 16.2 & 38.7  \\
             \ding{55} & \ding{51} & 17.7 & 38.7 \\
            \rowcolor{Gray}
             \ding{51} & \ding{51} & \textbf{18.7} & \textbf{38.7} \\
            \hline
        \end{tabular}
        \centering \caption{Ablation study on $W_t$.}
        \label{tab:ablation_w}
\end{table}

\begin{table}[h]
    \centering
    \small
    \begin{tabular}{c|cc|cc}
    \hline
    
    \multirow{2}{*}{\begin{tabular}[x]{@{}c@{}}\# basis \\ (b) \end{tabular}} &  \multicolumn{2}{c|}{ $|\mathcal{T}_\mathcal{E}|=9 (=3^2)$} & \multicolumn{2}{c}{ $|\mathcal{T}_\mathcal{E}|=16 (=4^2)$} \\
    \cline{2-5}
     & mR@100 & R@100 & mR@100 & R@100  \\ \hline
    4        & 17.2   & 38.5 & 17.2   & 38.5  \\
    \rowcolor{Gray}
    8        & 18.7   & 38.7 & 18.2   & 38.4  \\
    12       & 17.6   & 38.3 & 17.6   & 38.5  \\
    16       & 18.2   & 38.9 & 17.6   & 38.3  \\ \hline
    \end{tabular}
    
    \caption{Analysis on the number of basis matrices $b$.
    }
    \label{tab:ontol}
\end{table}

\subsubsection{Ablation Studies. }
\textbf{i) On the relation type-specific weight matrices: }
To verify the importance of capturing the semantics of relations in the edge-wise and update node-wise update, we remove the relation type-specific weight matrices in each update. More specifically, in the edge-wise step, we replace the $W_{\phi(u,v)}^{s2r}$ and $W_{\phi(u,v)}^{o2r}$ with $W^{s2r}$ and $W^{o2r}$, respectively, which implies that we treat all the relation types equivalently (i.e., Edge \ding{55}). Likewise, in the node-wise update step, we replace $W_{\phi(u,v)}^{r2s}$ and $W_{\phi(u,v)}^{r2o}$ with $W^{r2s}$ and $W^{r2o}$, respectively (i.e, Node \ding{55}). We have the following observations in Table~\ref{tab:ablation_w}: \textbf{1)} Adding the relation-specific weight matrix to either edge- and node-wise updates improves the overall performance of~\proposed. \textbf{2)} Considering the relations in both edge- and node-wise updates performs the best, which verifies the benefit of capturing relation-aware context for the SGG task.
\textbf{ii) On the efficiency of shared basis matrices: }
To verify the efficiency of composing a relation type-specific projection matrix as a linear combination of shared basis matrices, we evaluate~\proposed~over various number of basis (i.e., $b$) and relation types (i.e., $|\mathcal{T}_{\mathcal{E}}|$). 
Table \ref{tab:ontol} shows that even though $|\mathcal{T}_{\mathcal{E}}|$ increases from 9 to 16, the best performing number of basis matrices does not change, i.e., $b=8$ performs the best. This implies that RMP efficiently captures the semantics of relations even with a small number of parameters. Hence, we argue that a further benefit of adopting the basis matrices is that the complexity is expected to remain practical even if the number of relation types increases.

\begin{figure}[h]
    \centering
    \includegraphics[width=1.0\columnwidth]{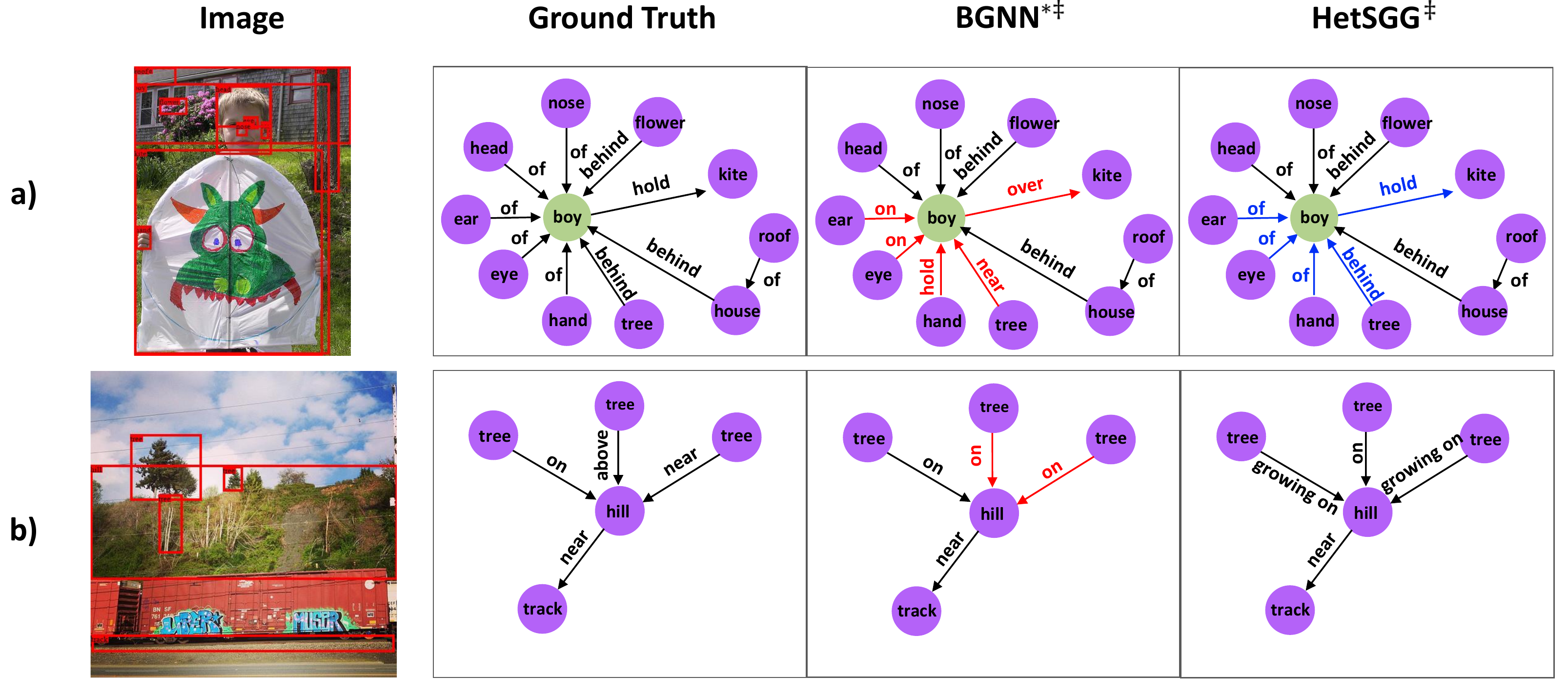}
    \vspace{-2.8ex}
    \caption{Qualitative comparisons between $\text{\proposed}^{\ddagger}$ and $\text{BGNN}^{\ast \ddagger}$ \cite{bgnn} in SGCls. 
    (\textbf{Red}: incorrect predictions by $\text{BGNN}^{\ast \ddagger}$,
    \textbf{Blue}: correct predictions by $\text{\proposed}^{\ddagger}$, but incorrect by BGNN.)
    }
    \label{fig:qualitative_result}
\end{figure}

\subsubsection{Qualitative Results. }
To verify that~\proposed~indeed captures the relation-aware context,
we qualitatively compare scene graphs generated by BGNN~\cite{bgnn} and~\proposed~for test images. We have the following observations:
\textbf{1)} Consider the triplet $\langle \textsf{hand}, \textsf{of}, \textsf{boy} \rangle$ in Figure~\ref{fig:qualitative_result}(a). We observe that BGNN generates an (incorrect) predicate, i.e., $\textsf{hold}$, even though it does not make sense for ``Product'' (\textsf{hand}) to \textsf{hold} ``Human'' (\textsf{boy}). On the other hand, since~\proposed~considers the object types, it can avoid such irrational cases, and generate the correct predicate, i.e., $\textsf{of}$.
\textbf{2)} Consider the triplet $\langle \textsf{tree}, \textsf{on}, \textsf{hill} \rangle$ in Figure~\ref{fig:qualitative_result}(b). Although BGNN generates a correct prediction, i.e., $\textsf{on}$, it simply predicts all the predicates related to \textsf{tree} as $\textsf{on}$, which is the most frequently appearing predicate in the dataset.
It is interesting to see that the prediction generated by~\proposed, i.e., \textsf{{growing on}}, though incorrect, is in fact more realistic. 
This again verifies that~\proposed~generates predicates that appear less frequently in the dataset.

\begin{table}[t]
\centering
\small
\begin{tabular}{c@{~}|c@{~}c|c@{~}c|c@{~}}
\hline
\textbf{Model} & \textbf{mR@50} & \textbf{R@50} & $\textbf{wmAP}_\text{rel}$ & $\textbf{wmAP}_\text{phr}$ & $\textbf{score}_\text{wtd}$ \\ \hline \hline

RelDN                  & 37.2                           & 75.3                          & 32.2            & 33.4           & 42.0                       \\
VCTree                 & 33.9                           & 74.1                          & 34.2            & 33.1           & 40.2                       \\
G-RCNN                 & 34.0                           & 74.5                          & 33.2            & 34.2           & 41.8                       \\
Motifs                 & 32.7                           & 71.6                          & 29.9            & 31.6           & 38.9                       \\
Unbiased               & 35.5                           & 69.3                          & 30.7            & 32.8           & 39.3                       \\
GPS-Net                & 38.9                           & 74.7                          & 32.8            & 33.9           & 41.6                       \\
$\text{BGNN}^{\ddagger}$                   & 40.5                           & 75.0                       & 33.5            & 34.1           & 42.1               \\ \cline{1-6} 
$\text{\proposed}^{\ddagger}$                       & 42.7                  & \textbf{76.8}                 & \textbf{34.6}   & \textbf{35.5}   & \textbf{43.3}                       \\
$\text{\proposed}^{\ddagger}_\texttt{++}$                       & \textbf{43.2}                  & 74.8                 & 33.5   & 34.5   & 42.2   \\ \hline
\end{tabular}
\caption{Results on Open Images V6 in SGGen task.}
\label{table:open_images}
\end{table}

\subsection{Open Images}
\label{sec:oi_v6}

\subsubsection{Dataset Details.} We closely follow the data processing and evaluation protocols of previous works~\cite{kuznetsova2020open,gps_net}. After preprocessing, OI V6 has 301 object classes, and 31 predicate classes, and is split into 126,368 train images, 1,813 validation images, and 6,322 test images.

\subsubsection{Comparisons with State-of-the-Art Methods. }
Table~\ref{table:open_images} demonstrates the experimental results in SGGen task on Open Images V6. 
We have the following observations:
\textbf{1)}~\proposed~and~$\proposed_{\texttt{++}}$~show large improvements on mR@50 implying that the bias prediction problem is greatly alleviated by our proposed framework.
\textbf{2)}~\proposed~and~$\proposed_{\texttt{++}}$ show competitive performance on R@50 and weighted mAP (wmAP). Considering that R@50 and wmAP are metrics that contradict with the goal of addressing the long-tail problem, this implies that~\proposed~alleviates the biased prediction problem, while also maintaining the performance on head classes.

\section{Conclusion}
In this work, we proposed an \textit{unbiased} \textit{heterogeneous} scene graph generation framework, called~\proposed. We devised a novel MPNN architecture, called relation-aware message passing network (RMP), that captures the relation-aware context given the types of objects and associated predicates. 
By considering a scene graph as a heterogeneous graph,~\proposed~alleviated the biased prediction problem incurred by the long-tail predicate class distribution.
\proposed~is model-agnostic in that it can be adopted to any MPNN-based SGG methods.
Through extensive experiments, we verified that~\proposed~outperforms existing SGG methods and generates more realistic scene graphs. 

\section*{Acknowledgements}
This work was supported by Institute of Information \& Communications Technology Planning \& Evaluation (IITP) grant funded by the Korean government (MSIT) (No. 2020-0-00004, Development of Previsional Intelligence based on Long-term Visual Memory Network, and No.2022-0-00077).

\footnotesize

\nocite{*}

\bibliography{aaai23.bib}

\clearpage
\appendix

\section{RMP adopted to BGNN \cite{bgnn} (i.e., \proposed\texttt{++})}
\label{appendix:bgnn}
As mentioned in the main paper, since RMP is model-agnostic, it can be adopted to any MPNN-based SGG methods, such as Graph R-CNN~\cite{graph_rcnn} and BGNN~\cite{bgnn}. In the main paper, we only explained RMP adopted to Graph R-CNN (i.e.,~\proposed), which is the simplest MPNN, due to the space limitation. Here, we describe~\proposed\texttt{++}, which is an RMP adopted to the adaptive message passing (AMP) of BGNN. More precisely, AMP propagates messages in a pairwise manner, that is, the messages are computed from the interaction between an object and only one of its neighboring objects without accounting for all the neighboring objects.
Hence, the main difference of~\proposed\texttt{++} compared with~\proposed~lies in the way that the importance of an object is computed in the edge-wise and node-wise update steps.

For the edge-wise update step of~\proposed\texttt{++}, the relation representation is computed as follows:
\vspace{-1ex}
\begin{equation}
\small
    z_{u\rightarrow v}^{(l+1)} = z_{u\rightarrow v}^{(l)} + \sigma ( \alpha_{\textsf{s2r}}(u) W_{\psi(u,v)}^{\textsf{s2r}} z_u^{(l)} + \alpha_{\textsf{o2r}}(v) W_{\psi(u,v)}^{\textsf{o2r}} z_v^{(l)})
    \label{eqn:1}
\end{equation}
\begin{equation}
\small
    \alpha_{\textsf{s2r}}(u) = \text{sigmoid}(w_{\textsf{s2r}}^{T} z_{u}^{(l)}), \quad  \alpha_{\textsf{o2r}}(v) = \text{sigmoid}(w_{\textsf{o2r}}^{T} z_{v}^{(l)})
    \label{eqn:2}
\end{equation}
where $z_{u\rightarrow v}^{(l+1)}\in\mathbb{R}^d$ is the relation representation of $e_{u\rightarrow v}$ at the $(l+1)$-th layer, $\sigma$ is a non-linear activation function, $W_{\psi(u,v)}^{\textsf{s2r}}\in\mathbb{R}^{d\times d}$ and $W_{\psi(u,v)}^{\textsf{o2r}}\in\mathbb{R}^{d\times d}$ are weight matrices of the relation type $\psi(u,v)$ for the two-way messages (i.e., \textsf{subject}-to-\textsf{relation} and \textsf{object}-to-\textsf{relation} messages given a triplet $\langle$\textsf{subject}, \textsf{predicate}, \textsf{object}$\rangle$), and $w_{\textsf{s2r}}\in \mathbb{R}^d$ and $w_\textsf{o2r} \in \mathbb{R}^d$ are attention vectors for \textsf{subject-to-relation} updates and \textsf{object-to-relation} updates. The main difference with the edge-wise update step of~\proposed~is that in~\proposed\texttt{++}, the importance of the messages propagated from \textsf{subject} (i.e., $u$) and \textsf{object} (i.e., $v$) is computed independently.
In summary, Equation 1 in the main paper is computed by Equations~\ref{eqn:1} and~\ref{eqn:2}~above.

For the node-wise update step of~\proposed\texttt{++}, the importance of object $v$ with regard to object $u$ is computed as follows:
\begin{equation}
\scriptsize
    \alpha_{\textsf{r2s}}(v,t) = \text{sigmoid}(w_{\textsf{r2s,t}}^{T} z_{u\rightarrow v}^{(l+1)}), \quad  \alpha_{\textsf{r2o}}(v,t) = \text{sigmoid}(w_{\textsf{r2o,t}}^{T} z_{v\rightarrow u}^{(l+1)})
    \label{eqn:3}
\end{equation}
where $w_{\textsf{r2s,t}}\in \mathbb{R}^d$ and $w_\textsf{r2o,t} \in \mathbb{R}^d$ are attention vectors for \textsf{relation-to-subject} updates and \textsf{relation-to-object} updates considering the relation type $t$, respectively. 
The main difference with the node-wise update step of~\proposed~is that in~\proposed\texttt{++}, the importance of object $v$ with regard to object $u$ is computed regardless of the neighboring objects of object $u$.
In other words, while~\proposed~determines the importance of messages among the neighboring objects,~\proposed\texttt{++} determines the importance of messages by only considering each node pair independently.

\begin{table*}[t]
\centering
\resizebox{0.85\textwidth}{!}{
\begin{tabular}{c|cc|c|cc|c|cc|c}
\hline
\multirow{2}{*}{\textbf{Models}} & \multicolumn{2}{c|}{\textbf{PredCls}}  &  \multirow{2}{*}{\textsf{Mean}} & \multicolumn{2}{c|}{\textbf{SGCls}}    & \multirow{2}{*}{\textsf{Mean}} & \multicolumn{2}{c|}{\textbf{SGGen}}  & \multirow{2}{*}{\textsf{Mean}}   \\ \cline{2-3} \cline{5-6} \cline{8-9}
                                 & \textbf{mR@50/100} & \textbf{R@50/100} & & \textbf{mR@50/100} & \textbf{R@50/100} & &\textbf{mR@50/100} & \textbf{R@50/100} & \\ \hline
PCPL \cite{yan2020pcpl}                            & 35.2/37.8          & 50.8/52.6         &  44.1
&  18.6/19.6            & 27.6/28.4        &  23.6
& 9.5/11.7            & 14.6/18.6        &  13.6 \\
DT2 \cite{Desai_2021_ICCV}                            & 35.9/39.7          & 23.3/25.6         &  31.1
& 24.8/27.5           & 16.2/17.6        & 21.5
& 22.0/24.4           & 15.0/16.3        &  19.4 \\
RTPB \cite{chen2022resistance}                            & 36.2/38.1          & 45.6/47.5         &  41.9
& 21.8/22.8           & 24.5/25.5        & 23.7
& 16.5/19.0            & 19.7/23.4        &  19.7 \\

DLFE\cite{DLFE} & 25.3/27.1 & 51.8/53.5 &39.4 &18.9/20.0 &33.5/34.6 & 26.8 & 11.8/13.8 & 22.7/26.3 & 18.7 \\

PPDL\cite{PPDL} & 32.3/33.3 & 47.2/47.6 & 40.1 & 17.5/18.2 & 28.4/29.3 & 23.4 & 11.4/13.5 & 21.2/23.9 & 17.5 \\

NICE-Motif\cite{NICE} & 29.9/32.3 & 55.1/57.2 & 43.6 & 16.6/17.9 & 33.1/34.0 & 25.4 & 12.2/14.4 & 27.8/31.8 & 21.6 \\

SHA\cite{dong2022stacked} & 41.6/44.1 & 35.1/37.2 & 39.5 & 23.0/24.3 & 22.8/23.9 & 23.5 & 17.9/20.9 & 14.9/18.2 & 18.0 \\

$\text{BGNN}^{\ddagger}$ \cite{bgnn}                            & 29.7/31.7          & 57.8/60.0         &  44.8
& 14.6/16.0           & 36.9/38.1        & 26.4
& 10.9/13.1            & 30.2/34.9       &  22.3\\ \hline
$\text{\proposed}^{\ddagger}$                           & 31.6/33.5          & 57.8/59.1     & 45.5
& 17.2/18.7          & 37.6/38.7    & \textbf{28.1}
& 12.2/14.4          & 30.0/34.6  & \textbf{22.8}       \\ 
$\text{\proposed}^{\ddagger}\texttt{++}$                           & 32.3/34.5     & 57.1/59.4         &      \textbf{45.8}
& 15.8/17.7  & 37.6/38.5    & 27.4
& 11.5/13.5          & 30.2/34.5    & 22.4    \\
\hline
\end{tabular}
}
\caption{Results on Visual Genome in terms of mR@50/100 and R@50/100 for PredCls, SGCls, and SGGen tasks. This table compares~\proposed~and~\proposed\texttt{++} with the state-of-the-arts that focus on solving the long-tail problem. \textsf{Mean} denotes the average of four values, i.e., mR@50/100 and R@50/100, which shows the overall performance of each method.}
\label{table:comparison}
\vspace{-2ex}
\end{table*}

\section{Dataset Details.}
\label{appendix:details}

\noindent\textbf{Visual Genome} 
We follow the same pre-processing strategies that have been widely used for evaluations of SGG \cite{IMP,neural_motif,bgnn}. Specifically, the most frequently appearing 150 object classes and 50 predicate classes are used for evaluation.
After preprocessing, each image contains 11.6 objects and 6.2 predicates on average.
A total of 108k images are split into training set (70\%) and test set (30\%), where 5k images from the training set are used for the model validation. 

\label{sec:oi}
\noindent \textbf{Open Images} 
Open Images V4/V6 is a large-scale data with the high-quality annotations recently proposed by Google~\cite{kuznetsova2020open}. We closely follow the data processing and evaluation protocols of previous works~\cite{kuznetsova2020open,gps_net,bgnn}. After preprocessing,
Open Images V6 has 301 object classes, and 31 predicate classes, and is split into 126,368 train images, 1,813 validation images, and 6,322 test images. Moreover, Open Images V4 in which we show the experiment result in Appendix~\ref{sec:all_oi_v4} has 57 object classes, and 9 predicate classes, and is split into 53,953 train images and 3,234 validation images.

\section{Additional Experiments on Visual Genome}
\label{sec:3}
 \subsection{Comparisons with other State-of-the-Arts.}

In addition to the compared methods in the main paper, there are other methods that focus on solving the long-tailed problem in SGG, such as DT2 \cite{Desai_2021_ICCV}, PCPL \cite{yan2020pcpl}, RTPB~\cite{chen2022resistance}, SHA \cite{dong2022stacked}, and DLFE \cite{DLFE}. These methods mainly alleviate the bias of the trained model with several strategies, showing significant improvements on the bias-sensitive metric, i.e., mean Recall. Detailed descriptions on these methods are provided in Appendix~\ref{sec:2}. 
Table~\ref{table:comparison} shows the performance of~\proposed~and~\proposed\texttt{++} compared with the performance of methods that focus on the long-tailed problem in SGG.
We have the following observations:
\textbf{1)} Although the compared unbiased SGG models greatly improve the performance of all three tasks in terms of mR@50/100, they perform poorly in terms of R@50/100. 
\textbf{2)} In contrast,~\proposed~and~\proposed\texttt{++} consistently show high performances in terms of both mR@50/100 and R@50/100,
which is shown by the superior performance in terms of an average of mR@50/100 and R@50/100.
From the above results, we argue that our proposed framework improves the overall performance without sacrificing the performance on the head predicates.

 \subsection{Performance Analysis for Each Class}
\begin{figure}[t]{
    \centering
    \includegraphics[width=0.9\columnwidth]{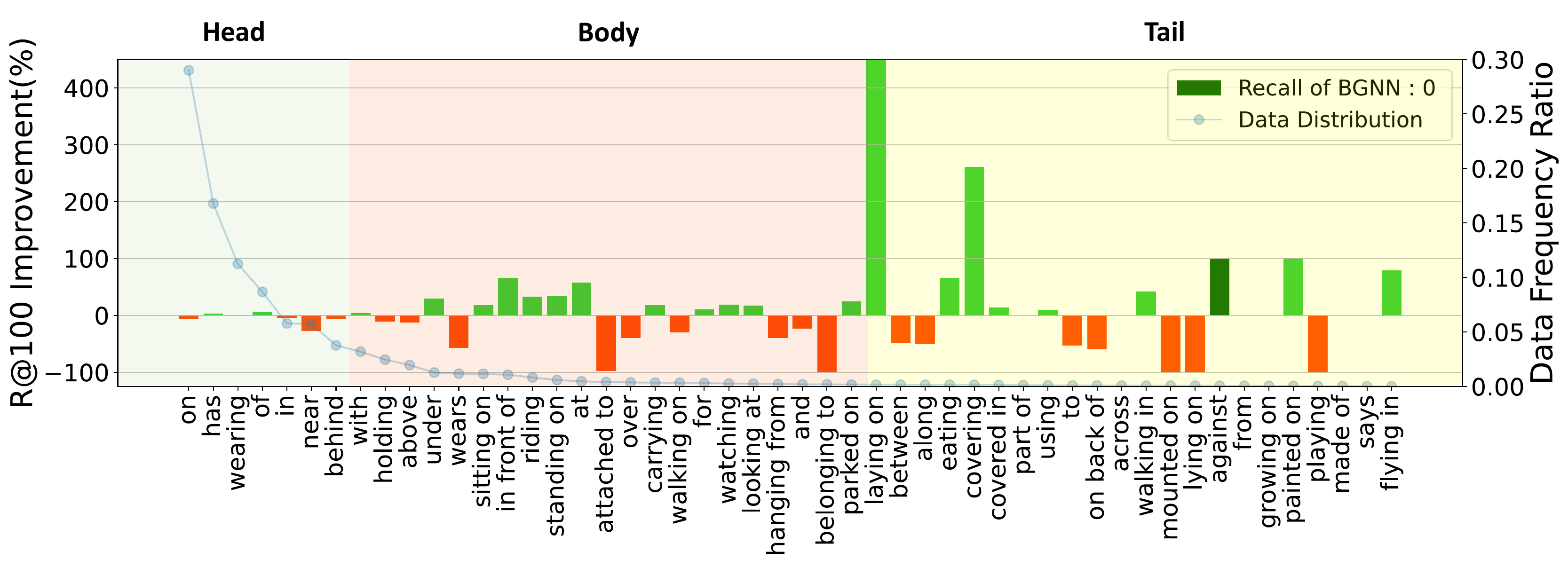}
    \vspace{-2ex}
    \caption{R@100 improvement per class of~$\text{{\proposed}}^{\ddagger}$~over $\text{BGNN}^{\ast \ddagger}$ in SGGen task.
    }
    \label{fig:per_class_sgdet}
}\end{figure}

\begin{figure}[t]{
    \centering
    \includegraphics[width=0.9\columnwidth]{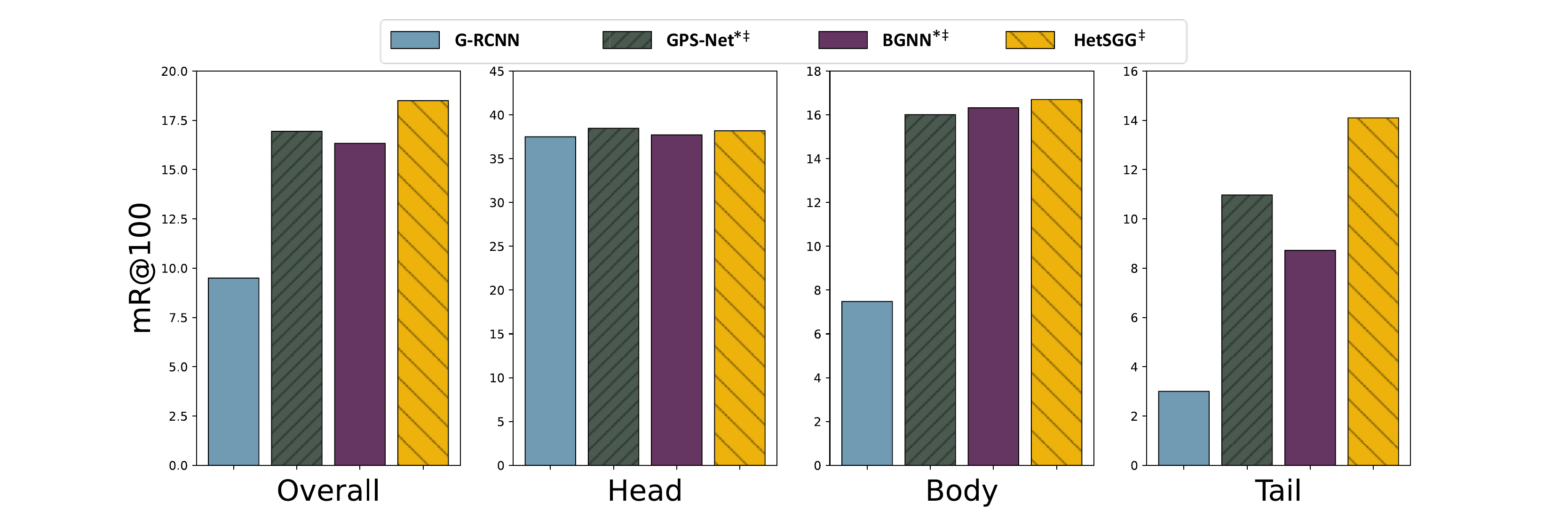}
    \caption{Results on the overall, head, body, and tail predicate classes in SGCls task.
    }
    \label{fig:per_class_sgcls_hbt}
}\end{figure}

In Figure \ref{fig:per_class_sgdet}, we compare the per class R@100 of~\proposed~with BGNN~\cite{bgnn}.
We observe that~\proposed~shows significant improvements in numerous body and tail predicate classes, while showing a competitive performance with BGNN in head classes.
In Figure~\ref{fig:per_class_sgcls_hbt}, we clearly see that~\proposed~shows a large improvement on the overall and tail classes, while showing a higher or similar performance with other baselines in head and body predicate classes.

\hfill
\subsection{Hyperparameter Sensitivity Analysis}
\label{appendix:hyper}
\begin{figure}[t]{
    \centering
    \includegraphics[width=0.98\columnwidth]{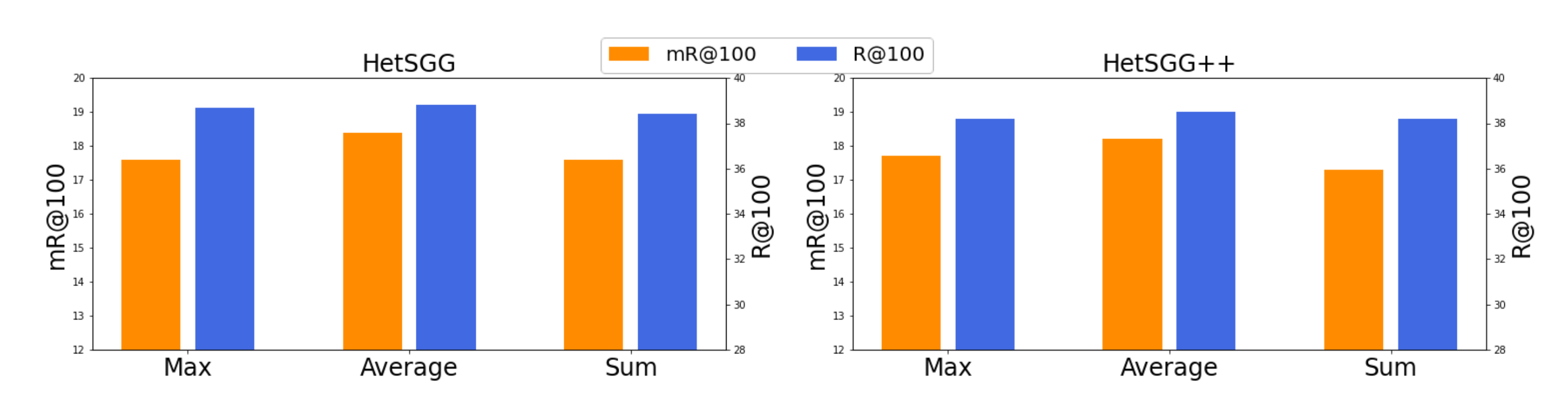}
    \caption{\textbf{The sensitivity of~$\text{{\proposed}}^{\ddagger}$ and ~$\text{{\proposed}}^{\ddagger}$\texttt{++} over aggregation functions for the type inference process}. $x$-axis: aggregation function. $y$-axis: mR@100, R@100 in SGCls task.
    }
    \label{fig:agg_hetsgg}
}\end{figure}

\begin{figure}[t]{
    \centering
    \includegraphics[width=0.9\columnwidth]{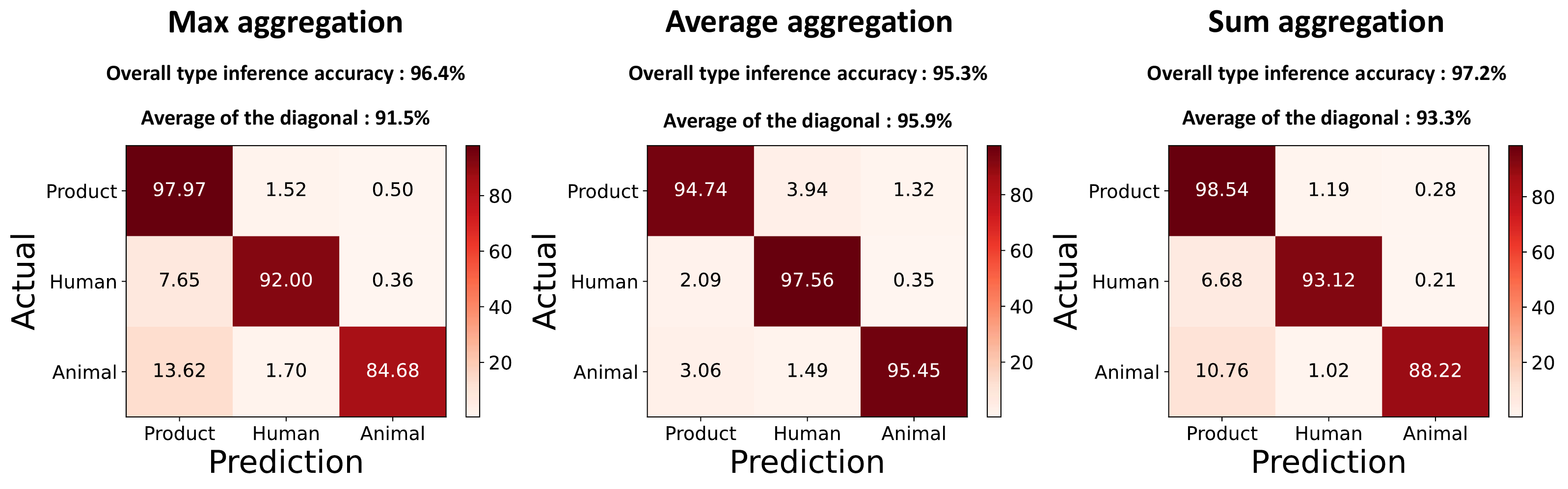}
    \caption{Normalized confusion matrix of object type inference for each aggregation function on VG.}
    \label{fig:confusion_matrix}
}\end{figure}

\begin{figure}[t]{
    \centering
    \includegraphics[width=0.95\columnwidth]{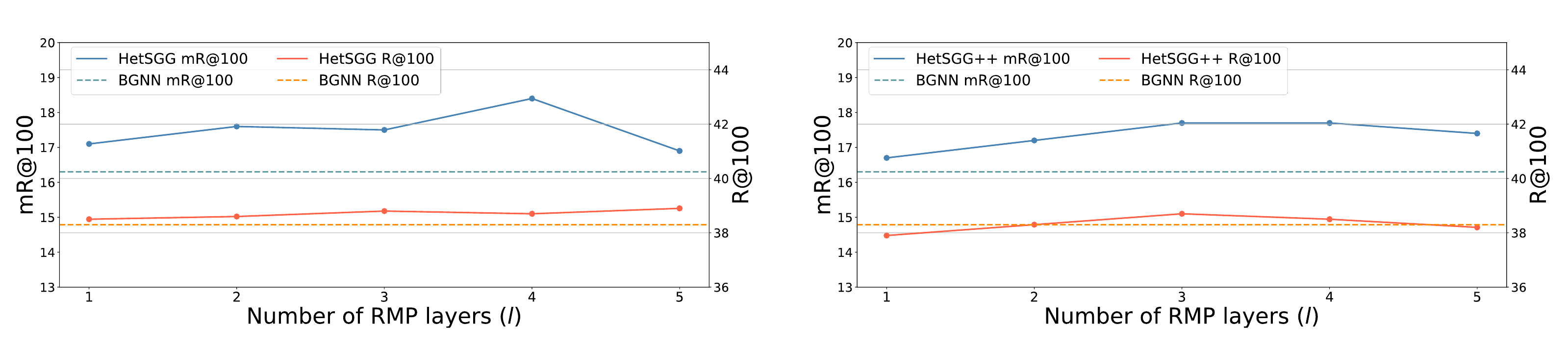}
    \caption{\textbf{The sensitivity of~$\text{{\proposed}}^{\ddagger}$ and~$\text{{\proposed}}^{\ddagger}\texttt{++}$ over the number of RMP layers}. $x$-axis: the number of RMP layers (\textit{l}). $y$-axis: mR@100 and R@100 in SGCls task.
    }
    \label{fig:nlayer_hetsgg}
}\end{figure}

We further analyze the sensitivity of~\proposed~over several hyperparameters. We mainly investigate the effect of \textbf{i)} the number of RMP layers (i.e., $l$) (Figure~\ref{fig:nlayer_hetsgg}), and \textbf{ii)} the aggregation functions for the type inference process (Figure~\ref{fig:agg_hetsgg} and~\ref{fig:confusion_matrix}).

\smallskip
\noindent\textbf{i) Number of layers.} In Figure \ref{fig:nlayer_hetsgg}, we observe that stacking more RMP layers improves the performance of both~\proposed~and~\proposed\texttt{++}, and the highest mR@100 is achieved with four layers. This implies that capturing high-order interactions between objects with multiple RMP layers is effective for learning the contextual representation of objects and predicates. On the other hand, stacking more than four layers degrades mR@100. We conjecture that this is mainly due to the over-smoothing problem of GNNs~\cite{li2018deeper}, implying that an appropriate number of layers should be found.
Moreover, we observe that~\proposed~and~\proposed\texttt{++} generally outperform BGNN regardless of the number of RMP layers, which demonstrates the superiority of RMP of~\proposed~over AMP of BGNN.

\smallskip  
\noindent\textbf{ii) Aggregation Function.} 
As shown in Table 2 of the main paper, the performance of~\proposed~is dependent on the accuracy of the object type inference process. 
Since the type inference process involves the selection of an aggregation function to aggregate the class logits, we evaluate the performance of~\proposed~over various aggregation functions used for the object type inference process. Figure~\ref{fig:agg_hetsgg} shows that among various aggregation functions, using \textsf{Average} performs the best.
Furthermore, we investigate how the object types (i.e., ``Product'', ``Animal'', and ``Human'') are actually assigned according to the aggregation functions in Figure~\ref{fig:confusion_matrix}.
We observe that \textsf{Sum} and \textsf{Max} show high accuracy in terms of the overall type assignment, achieving 97.2\% and 96.4\% respectively, whereas \textsf{Average} shows a lower accuracy of 95.3\%. 
However, we argue that the average performance of the diagonal entries of each matrix is the key to the success of~\proposed, as our ultimate goal is to accurately infer the predicate type rather than the object type. More precisely, the predicate type of an object pair $(u,v)$ is considered to be correctly determined, if the types of both objects (i.e., $u$ and $v$) are correctly assigned. In this regard, since \textsf{Average} achieves the highest diagonal accuracy (i.e., average of the diagonal), we choose \textsf{Average} as our choice of the aggregation function. 
 
\section{Additional Experiments on Open Images}

\subsection{Comparison In OI V4.}
\label{sec:all_oi_v4}
We not only compare with state-of-the-arts method in Open Images V6 shown in Table 5, but also in Open Images V4. As shown in Table~\ref{table:oiv4_main}, \proposed~and \proposed\texttt{++} unleash the mR@50 and R@50 performance, which demonstrates the effectiveness of alleviating the long-tailed predicate distribution. The~\proposed~shows the lower performance on $\text{wmAP}_{\text{rel}}$, $\text{wmAP}_{\text{phr}}$, and $\text{score}_{\text{wtd}}$ while they outperform the performance on Open Images V6. We attribute the results to the fact that it only contains 9 predicate classes, which makes it relatively easier to classify compared with Open Images V6 that contains 31
predicate classes. We argue that ~\proposed~performs well
when more fine-grained and complex relations are given,
and thus we expect it to perform well as the number of predicate classes increases demonstrating the practicality of ~\proposed. Moreover, weighted mAP (wmAP) is a metric that contradicts with the goal
of addressing the long-tail problem, which we nevertheless
reported for comprehensiveness of the experiments.

\subsection{Analysis of $\text{AP}_{\text{rel}}$ per class on OI V6.}
\begin{figure}[t]
{
    \centering
    \includegraphics[width=0.9\columnwidth]{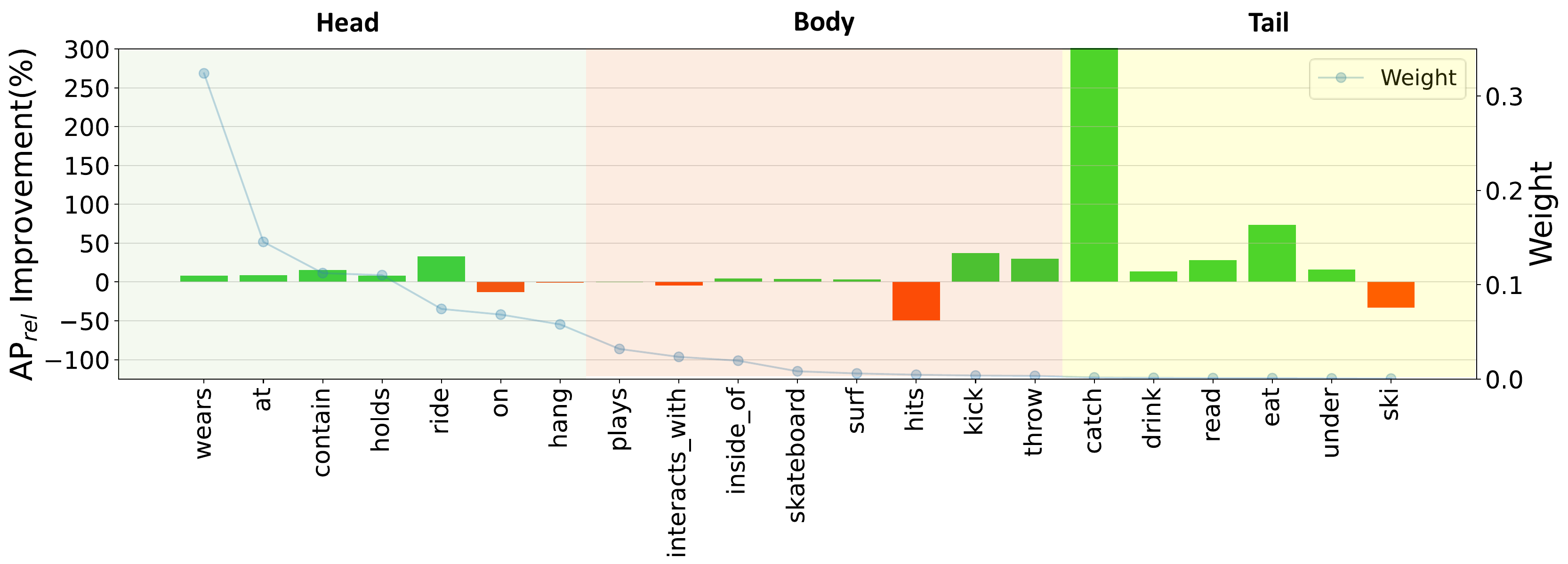}
    \caption{AP$_{rel}$ improvement per class of~\proposed~compared with BGNN on Open Images V6. \textit{Weight} (i.e., right $y$-axis) denotes the frequency ratio in the test data. Note that wmAP$_\text{rel}$ is computed by the sum of \textit{weight} $\times$ AP$_\text{rel}$ of all predicates.}
    \label{fig:ap_per_oi}
}
\end{figure}

For more detailed analysis on Open Images V6, we compare the per class $\text{AP}_{\textsf{rel}}$ of~\proposed~and that of BGNN \cite{bgnn}.
Figure \ref{fig:ap_per_oi} shows the per class improvement of~\proposed~in terms of AP$_{\text{rel}}$. Among 30 predicate classes in Open Images V6, 9 predicate classes are excluded in Figure~\ref{fig:ap_per_oi} as they do not appear in the test data. The order of predicate classes of $x$-axis is sorted by the predicate frequency, i.e., \textsf{wears} is the most frequent class and \textsf{ski} is the most rare class in the predicate class distribution of the test data. We observe that~\proposed~generally outperforms BGNN in terms of $\text{AP}_{\text{rel}}$ not only on tail predicate classes, but also on head classes. Consequently,~\proposed~outperforms BGNN on $\text{wmAP}_{\textsf{rel}}$ and $\text{score}_{\textsf{wtd}}$ as shown in Table 5 of the main paper.

\begin{table}[t]
\resizebox{.95\columnwidth}{!}{
\begin{tabular}{c|cc|cc|c}
\hline
\multirow{2}{*}{\textbf{Model}} & \multirow{2}{*}{\textbf{mR@50}} & \multirow{2}{*}{\textbf{R@50}} & \multirow{2}{*}{$\textbf{wmAP}_\text{rel}$} & \multirow{2}{*}{$\textbf{wmAP}_\text{phr}$} & \multirow{2}{*}{$\textbf{score}_\text{wtd}$} \\
       &                                 &                                &      &     &                                      \\ \hline
RelDN \cite{RelDN}               & 70.40                           & 75.66                          & 36.13            & 39.91           & 45.21                                \\
GPS-Net \cite{gps_net}      & 69.50                           & 74.65                          & 35.02            & 39.40           & 44.70    \\
$\text{BGNN}^{\ddagger}$\cite{bgnn}       & 72.11                           & 75.46                          & \textbf{37.76}   & \textbf{41.70}  & \textbf{46.87}                       \\ \cline{1-6} $\text{\textbf{\proposed}}^{\ddagger}$                       & \textbf{77.22}                  & \textbf{80.40}                 & 36.11            & 40.14           & 45.79                                \\
$\text{\textbf{\proposed}}^{\ddagger}\texttt{++}$                       & 75.83                  & 79.05                 & 36.53   & 40.40   & 45.91            \\ \hline
                             
\end{tabular}
}
\caption{Results on Open Images V4 in SGGen task.}
\label{table:oiv4_main}
\end{table}

\section{Detailed Descriptions of Baselines}
\label{sec:2}
Compared baselines include models using  1) message passing frameworks to learn the context \cite{msdn,graph_rcnn,gps_net,bgnn} (\textbf{MPNN-based models}), 2) strategies to alleviate the long-tail problem \cite{unbiased,Desai_2021_ICCV,chen2022resistance,yan2020pcpl,NICE,dong2022stacked,DLFE,PPDL} (\textbf{Unbiased models}), and 3) various architectures to learn a flexible representation \cite{RelDN,neural_motif,VCTree} (\textbf{Others}).

\begin{itemize}
    \item \textbf{MPNN-based models}
    
    \smallskip
    \begin{itemize}
        \item MSDN \cite{msdn} is a scene graph generation method that simultaneously performs the image captioning task. It constructs a dynamic graph and utilizes a message passing network to capture the context.
        \item Graph R-CNN \cite{graph_rcnn} utilizes a relation proposal network and the attentive message propagation to recognize important visual context and relations. 
        \item GPS-Net~\cite{gps_net} uses several properties in SGG including the direction-aware MPNNs to capture the direction-aware context. 
        \item BGNN~\cite{bgnn} employs the adaptive message passing scheme to control the noise in a scene graph. Based on the novel re-sampling strategy, i.e., bi-level sampling, BGNN is a powerful framework which achieves both the state-of-art performance. 
    \end{itemize}
    
    \item \textbf{Unbiased models} 
    
    \smallskip
    \begin{itemize}
        \item Unbiased~\cite{unbiased} adopts causal inference in the prediction stage to make unbiased predictions.
        \item DT2 \cite{Desai_2021_ICCV} is a transfer learning-based approach that alleviates the long-tail problems inherent in both objects and predicates. 
        \item RTPB \cite{chen2022resistance} designs the training schemes that utilize existing predicate bias in SGG.
        \item PCPL \cite{yan2020pcpl} utilizes implicit correlation among predicate classes, and captures the contextual information by stacking the graph encoding module using attention.
        \item NICE \cite{NICE} regards the SGG problem as noisy label learning that solves out-of-distribution (OOD) problem. 
        \item SHA \cite{dong2022stacked} is a transfer learning-based approach that learns knowledges from several balanced groups. By utilizing the shared knowledge, SHA generates unbiased scene graphs.
        \item DLFE \cite{DLFE} considers the SGG problem as Positive Unlabeled (PU) learning. Based on the PU learning formulation, it alleviates the reporting biased problems that conspicuous classes are less predicted.
        \item PPDL \cite{PPDL} presents the biased training loss to re-balance the similarity between predicted predicate distribution.
    \end{itemize}
    
    \item \textbf{Others} 
    
    \smallskip
    \begin{itemize}
        \item RelDN \cite{RelDN} introduces a contrastive loss to solve the entity confusion and proximal relationship ambiguity.
        \item Motifs \cite{neural_motif} is the model that uses Bi-LSTM to capture the contextual information of objects in an image. 
        \item VCTree \cite{VCTree} proposes a binary tree structure that uses bidirectional-TreeLSTM to encode the visual context. 
    \end{itemize}
\end{itemize}

\section{Hyperparameter Configurations}
\label{appendix:hyper_config}
Table~\ref{table:hyper} presents the hyperparameter configurations to reproduce~\proposed~and~\proposed\texttt{++}. Note that we adopt the same configurations for all the tasks, i.e., PredCls, SGCls, and SGGen.

\begin{table}[h]
\centering

\resizebox{.98\columnwidth}{!}{%
\begin{tabular}{c|c|ccccc}
\hline
\multirow{3}{*}{\textbf{Model}}     & \multirow{3}{*}{\textbf{Dataset}} & \multicolumn{5}{c}{\textbf{Hyperparameters}}                                                                                                                                                                                                                                                                                                                                                                     \\ \cline{3-7} 
                                    &                                    & \multicolumn{1}{c|}{\multirow{2}{*}{\begin{tabular}[c]{@{}c@{}}\textbf{Num.}\\ \textbf{Bases($b$)}\end{tabular}}} & \multicolumn{1}{c|}{\multirow{2}{*}{\begin{tabular}[c]{@{}c@{}}\textbf{Dimension}\\  \textbf{size ($d$)}\end{tabular}}} & \multicolumn{1}{c|}{\multirow{2}{*}{\begin{tabular} [c]{@{}c@{}}\textbf{Num.}\\
                                    \textbf{Layers($l$)}\end{tabular}}}
                                    & \multicolumn{1}{c|}{\multirow{2}{*}{\begin{tabular}[c]{@{}c@{}}\textbf{Num. rel}\\  \textbf{types ($|\mathcal{T}_{\mathcal{E}}|$)}\end{tabular}}} & \multicolumn{1}{c}{\multirow{2}{*}{\textbf{\begin{tabular}[c]{@{}c@{}}\textbf{Aggregation}\\  function\end{tabular}}}} \\
                                    &                                    & \multicolumn{1}{c|}{}                                                                        & \multicolumn{1}{c|}{}                                                                                  & \multicolumn{1}{c|}{}                       & \multicolumn{1}{c|}{}                                                                          & \multicolumn{1}{c}{}                             \\ \hline
\multirow{2}{*}{\textbf{\proposed}} & \textbf{Visual Genome}                      & \multicolumn{1}{c|}{8}                                                                       & \multicolumn{1}{c|}{128}                                                                               & \multicolumn{1}{c|}{4}                      & \multicolumn{1}{c|}{9}                                                                         & \textsf{Average($\cdot$)}                                       \\
                                    & \textbf{Open Images V6}                  & \multicolumn{1}{c|}{4}                                                                       & \multicolumn{1}{c|}{128}                                                                               & \multicolumn{1}{c|}{4}                      & \multicolumn{1}{c|}{9}                                                                         & \textsf{Average($\cdot$)}                                        \\ \hline
\multirow{2}{*}{\textbf{\proposed\texttt{++}}}    & \textbf{Visual Genome}                      & \multicolumn{1}{c|}{8}                                                                       & \multicolumn{1}{c|}{128}                                                                               & \multicolumn{1}{c|}{3}                      & \multicolumn{1}{c|}{9}                                                                         & \textsf{Average($\cdot$)}                                        \\
                                    & \textbf{Open Images V6}                  & \multicolumn{1}{c|}{4}                                                                       & \multicolumn{1}{c|}{128}                                                                               & \multicolumn{1}{c|}{3}                      & \multicolumn{1}{c|}{9}                                                                         & \textsf{Average($\cdot$)}                                        \\ \hline
\end{tabular}
}
\caption{Hyperparameter configurations on each dataset.}
\label{table:hyper}
\end{table}

\section{Object Type Assignments}
\label{sec:5}
In Table~\ref{table:type_vg}, we report how the object classes are mapped to object types in VG and Open Images V4/V6.

\begin{table}[t]

\resizebox{\columnwidth}{!}{
\begin{tabular}{c|c|l}
\hline \hline
\textbf{Dataset} & \textbf{Object Type} & \multicolumn{1}{c}{\textbf{Object Class}}                                                                                 \\ \hline
\multirow{3}{*}{\textbf{Visual Genome}} & \textbf{Product}     & \multicolumn{1}{l}{\begin{tabular}[c]{@{}l@{}}airplane, bag, banana, basket, \textcolor{blue}{beach}, bed, bench, bike, board, boat, book, boot, \\bottle, bowl, box, branch, \textcolor{blue}{building}, bus, cabinet, cap, car, chair, clock, coat, \\\textcolor{blue}{counter}, cup, curtain, desk, door, drawer, engine, \textcolor{blue}{fence}, flag, flower, food, fork,\\ fruit, glass, glove, hair, handle, hat, helmet, \textcolor{blue}{hill}, \textcolor{blue}{house}, jacket, jean, kite, lamp,\\ laptop, leaf, letter, light, logo, motorcycle, \textcolor{blue}{mountain}, number, orange, pant, \\ paper, phone, pillow, pizza, plane, plant, plate, \textcolor{blue}{pole}, post, pot, racket, \textcolor{blue}{railing}, \\rock, \textcolor{blue}{roof}, \textcolor{blue}{room}, screen, seat, \textcolor{blue}{shelf}, shirt, shoe, \textcolor{blue}{sidewalk}, sign, sink, skateboard, \\ski, sneaker, snow, sock, \textcolor{blue}{stand}, \textcolor{blue}{street}, surfboard, table, tie, tile, tire, \textcolor{blue}{toilet}, towel,\\\textcolor{blue}{tower}, \textcolor{blue}{track}, train, tree, truck, trunk, umbrella, vase, vegetable, vehicle, wave, \\wheel, window, windshield, wing, wire, arm, ear, eye, face, finger, hand, head, leg, \\mouth, neck, nose, paw, short, tail\end{tabular}} \\ \cline{2-3}
& \textbf{Human}       & boy, child, girl, guy, kid, lady, man, men, people, person, player, skier, woman\\ \cline{2-3}
& \textbf{Animal}      & animal, bear, bird, cat, cow, dog, elephant, giraffe, horse, sheep, zebra \\ \hline \hline

\multirow{3}{*}{\textbf{Open Image V4}} & \textbf{Product} & \begin{tabular}[c]{@{}l@{}}Piano, Tennis ball, Van, Football, Beer, Camera, Suitcase, Bench, Motorcycle, Mug, \\Tennis racket, Drum, Spoon, Surfboard, Bicycle, Knife, Rugby ball, Handbag, \\Microwave oven, Flute, Taxi Wine glass, Backpack, Racket, Table, Pretzel, Bed, \\Snowboard, Car, Chair, Microphone, Coffee cup, Table tennis racket, Bottle, Guitar, \\Desk,Ski, Coffee table, Chopsticks, Mobile phone, Sofa bed, Violin, Fork, Oven, Briefcase\end{tabular} \\ \cline{2-3}
& \textbf{Human}      & Boy, Man, Woman, Girl \\ \cline{2-3}
& \textbf{Animal}     & Dolphin, Horse, Hamster, Dog, Cat, Elephant, Monkey, Snake \\ \hline\hline       

\multirow{3}{*}{\textbf{Open Image V6}} & \textbf{Product} & \begin{tabular}[c]{@{}l@{}}football, ladder, organ (musical instrument), apple, paddle, beer, chopsticks, \\
croissant, cucumber, radish, doll, washing machine, belt, sunglasses, banjo, cart, \\
backpack, bicycle, boat, surfboard, boot, bus, bicycle wheel, waffle, pancake, \\
pretzel, bagel, teapot, bow and arrow, popcorn, burrito, balloon, tent, lantern, \\
tiara, limousine, necklace, scissors, chair, cheese, earrings, suitcase, muffin, \\
snowmobile, cello, jet ski, desk, juice, gondola, cannon, cookie, cocktail, box, \\
christmas tree, cowboy hat, hiking equipment, studio couch, drum, zucchini, oven, \\
cricket ball, whiteboard, fedora, scarf, sombrero, tin can, mug, stretcher, goggles, \\
roller skates, coffee cup, cutting board, volleyball (ball), coffee, whisk, sun hat, \\
tree house, flying disc, french fries, barrel, kite, tart, treadmill, french horn, \\
golf cart, egg (food), guitar, grape, houseplant, baseball bat, baseball glove, wheelchair, \\
stationary bicycle, hammer, sofa bed, adhesive tape, harp, sandal, bicycle helmet, \\
saucer, harpsichord, bed, drinking straw, indoor rower, punching bag, common fig, \\
golf ball, artichoke, table, knife, bottle, lynx, lavender (plant), dumbbell, bowl, \\
billiard table, motorcycle, swim cap, frying pan, snowplow, milk, plate, mobile phone, \\
mixing bowl, pitcher (container), personal flotation device, table tennis racket, \\
musical keyboard, briefcase, kitchen knife, tennis ball, plastic bag, oboe, piano, \\
potato, pasta, pumpkin, pear, infant bed, pizza, rifle, skateboard, high heels, rose, \\
saxophone, shotgun, submarine sandwich, snowboard, sword, sushi, loveseat, ski, \\
stethoscope, segway, coffee table, trombone, tea, tank, taco, torch, strawberry, trumpet, \\
tree, tomato, train, picnic basket, bowling equipment, football helmet, truck, violin, \\
handbag, wine, wok, jug, bread, helicopter, toilet paper, lemon, banana, wine glass, \\
countertop, tablet computer, waste container, book, axe, palm tree, hamburger, maple, \\
garden asparagus, airplane, spoon, oyster, horizontal bar, ice cream, parachute, orange, \\
closet, peach, coconut, fork, camera, racket, unicycle, cabbage, carrot, mango, flowerpot, \\
drawer, stool, cake, common sunflower, microwave oven, honeycomb, watch, candy, \\
salad, van, corded phone, tennis racket, serving tray, kitchen \& dining room table, \\
dog bed, cake stand, cat furniture, microphone, broccoli, grapefruit, bell pepper, lily, \\
pomegranate, doughnut, glasses, pen, car, teddy bear, watermelon, cantaloupe, flute, \\
balance beam, sewing machine, binoculars, rays and skates, accordion, taxi, canoe, \\
rugby ball, mushroom, candle, bench, platter, pineapple, handgun, crown, tripod\end{tabular} \\ \cline{2-3}
& \textbf{Human}      & boy, woman, man, girl\\ \cline{2-3}
& \textbf{Animal}     & 
\begin{tabular}[c]{@{}l@{}}
tortoise, sea turtle, brown bear, cat, dinosaur, dolphin, harbor seal, fox, panda, \\
horse, hamster, jaguar (animal), lizard, polar bear, snake, whale, monkey, alpaca\\
dog, elephant, shark, lobster, sea lion, skunk, crocodile, shrimp, crab, seahorse, \\
\end{tabular}
\\ \hline

\end{tabular}
}
\caption{Assignment of object classes to object types in VG and Open Image V4/V6. Note that \textcolor{blue}{\textbf{blue}} object classes in ``Product'' of Visual Genome denote classes that belong to ``Landform'' object type, which is required for experiments in Table 2 of the main paper.}

\label{table:type_vg}

\end{table}
\end{document}